\documentclass{article}

\PassOptionsToPackage{numbers, compress}{natbib}

\usepackage[preprint]{neurips_2024}

\setcitestyle{square} 
\usepackage[utf8]{inputenc} 
\usepackage[T1]{fontenc}    
\usepackage{hyperref}       
\usepackage{url}            
\usepackage{booktabs}       
\usepackage{amsfonts}       
\usepackage{nicefrac}       
\usepackage{microtype}      
\usepackage[[table,xcdraw]{xcolor}         
\usepackage{amsmath}
\usepackage{graphicx}
\usepackage{caption}
\usepackage{blindtext}
\newtheorem{definition}{Definition}[section]
 \usepackage{multirow}
 \usepackage{multicol}
 \usepackage{subcaption}

\hypersetup{
  colorlinks   = true, 
  urlcolor     = blue, 
  linkcolor    = red, 
  citecolor   = blue 
}

\title{ConceptPrune: Concept Editing in Diffusion Models via Skilled Neuron Pruning}

\newcommand\blfootnote[1]{%
  \begingroup
  \renewcommand\thefootnote{}\footnote{#1}%
  \addtocounter{footnote}{-1}%
  \endgroup
}

\author{Ruchika Chavhan$^1$, Da Li$^2$, Timothy Hospedales$^{1,2}$\\
$^1$University of Edinburgh, \\
$^2$Samsung AI Research Centre, Cambridge\\
}

\begin{document}

\maketitle

\begin{abstract}
While large-scale text-to-image diffusion models have demonstrated impressive image-generation capabilities, there are significant concerns about their potential misuse for generating unsafe content, violating copyright, and perpetuating societal biases. Recently, the text-to-image generation community has begun addressing these concerns by editing or unlearning undesired concepts from pre-trained models. However, these methods often involve data-intensive and inefficient fine-tuning or utilize various forms of token remapping, rendering them susceptible to adversarial jailbreaks. In this paper, we present a simple and effective training-free approach, \textbf{\textit{ConceptPrune}}, wherein we first identify critical regions within pre-trained models responsible for generating undesirable concepts, thereby facilitating straightforward concept unlearning via weight pruning. Experiments across a range of concepts including artistic styles, nudity, object erasure, and gender debiasing demonstrate that target concepts can be efficiently erased by pruning a tiny fraction, approximately 0.12\% of total weights, enabling multi-concept erasure and robustness against various white-box and black-box adversarial attacks. Our code is available at \url{https://github.com/ruchikachavhan/concept-prune.git}\blfootnote{Correspondence to: $\{$\texttt{ruchika.chavhan, t.hospedales}$\}$@ed.ac.uk}
\end{abstract}

\section{Introduction}
In recent years, text-to-image generation has witnessed significant advances driven by the development and adoption of diffusion models (DMs) \cite{ho2020denoising, rombach2021highresolution, ruiz2022dreambooth, saharia2022photorealistic, nichol2021glide, zhang2023survey, luo2023latent, podell2023sdxl} across industries and real-world scenarios. However, this swift advancement presents a substantial risk. Diffusion models can threaten artists' livelihoods through style replication \cite{courtcase}, generate convincing deepfakes and NSFW content \cite{mitreview, blog}, and perpetuate societal biases \cite{luccioni2023stable}. The risks associated with large-scale text-to-image models arise from billion-sized web-scraped datasets used in training, comprising public datasets like LAION \cite{schuhmann2022laionb}, COYO \cite{kakaobrain2022coyo-700m}, and CC12M \cite{changpinyo2021cc12m}, that often lack human-level quality assurance. A simplistic and naive solution to mitigate these risks involves fine-tuning the model on datasets without this undesired content; however, this approach can prove to be highly compute-expensive. 

Several efforts addressing the risks of diffusion models have been made from the perspective of Concept Editing \cite{kumari2023conceptablation, gandikota2023erasing, gandikota2023unified, zhang2023forgetmenot, orgad2023editing} and Model Unlearning (MU) \cite{heng2023selective, zhao2024separable, liu2024implicit, wu2024erasediff, fan2023salun}, both aimed at eliminating undesired prompts, albeit with differing objectives. Concept editing methods seek to eliminate undesired prompts by aligning latent representations of the target concept with a concept to be retained, via methods such as maximizing similarity \cite{kumari2023conceptablation, gandikota2023erasing} and token remapping \cite{zhang2023forgetmenot, gandikota2023unified}. Conversely, Model Unlearning formulates an objective that penalizes forgetting desired concepts while promoting the elimination of undesired ones, but this requires expensive computations and fine-tuning. Moreover, as most concept editing approaches rely on some form of token blacklisting or resteering \cite{zhang2023forgetmenot}, adversarial attacks based on textual inversion \cite{zhang2023generate, pham2023circumventing, yang2023mma, tsai2024ringabell} have demonstrated the ability to circumvent concept erasure methods \cite{gandikota2023erasing, gandikota2023unified, zhang2023forgetmenot} that were previously believed to be robust with a near-perfect success rate. 


In this paper, we introduce \textit{ConceptPrune}, an entirely training-free method for concept editing that, for the first time, tackles knowledge editing in diffusion models through the lens of pruning.  Leveraging recently introduced pruning heuristics \cite{sun2024simple}, we identify regions or neurons in feed-forward layers of diffusion models that strongly activate in the presence of a concept, and denote them as \textit{skilled neurons}. Subsequently, concept removal can be achieved by simply pruning or \textit{zeroing} out these skilled regions. We demonstrate that ConceptPrune provides a rapid, efficient, and unified solution for erasing undesired concepts, including various artist styles, nudity, undesired objects, and gender biases. Notably, it maintains the outstanding image-generation prowess of pre-trained models while remaining resilient to adversarial attacks. 


\section{Related Work}


\textbf{Diffusion model concept editing:} Most concept Editing works \cite{kumari2023conceptablation, gandikota2023erasing, gandikota2023unified, orgad2023editing, zhang2023forgetmenot} within diffusion models aim to eliminate target concepts by aligning the model's output with that of a reference prompt, whose concept we wish to retain. For example, to remove the concept `nudity', the target prompt can be formulated as \textit{``a photo of a naked person''} while the reference prompt can be \textit{``a photo of a person''}. Then the target concept ``nudity'' can be removed by minimizing certain metrics between denoised predictions of target and reference prompts \cite{kumari2023conceptablation}, utilizing score-based composition as unsupervised training data \cite{gandikota2023erasing}, or employing attention re-steering to reduce cross-attention weights for the target prompts~\cite{zhang2023forgetmenot}. Unlike approaches relying on latent representations, UCE \cite{gandikota2023unified} and MEMIT \cite{orgad2023editing} operate on token rewriting, adjusting attention module parameters in the UNet to align token embeddings corresponding to the target prompt with the reference prompt using a closed-form solution.

\textbf{Diffusion model unlearning (MU):} MU \cite{zhao2024separable, heng2023selective, fan2023salun, wu2024erasediff, liu2024implicit} operates with two separate datasets: the forgetting dataset and the retention dataset. 
The model is fine-tuned such that information from the forgetting dataset is erased while knowledge corresponding to the remaining data remains intact. There are different ways to achieve this dual objective optimization: a first-order dual problem formulation \cite{wu2024erasediff}, generative replay on the retention dataset to ensure consistent retention of the dataset \cite{heng2023selective}, and fine-tuning via saliency masks that retain the reference concept while disregarding the target concept \cite{fan2023salun}. While these methods have shown remarkable efficacy in unlearning multiple concepts, they are usually computationally expensive, especially for large-scale models.

In contrast to the existing Concept Editing and MU methods, our method operates on training-free neuron identification and pruning of critical regions that are responsible for generating undesired behaviors. While our method does not necessitate any compute-intensive fine-tuning of parameters, it is directly aligned with the Concept Editing line of work as we demonstrate that denoised prediction matching is possible through the selective removal of neurons in the weight space.


\textbf{Language model skilled neuron identification:} Previous works \cite{wang2022finding, suau2020finding, durrani2023discovering, dalvi2018grain, durrani2020analyzing, antverg2022pitfalls} present strong evidence that activation of specific neurons in feed-forward networks in transformers show high correlation with task labels, with perturbations to these neurons impacting task performance. Modular components within pre-trained transformers were identified by leveraging the inherent sparsity in neurons, as shown in \cite{zhang2022moefication}. Further, \cite{zhang2023emergent} demonstrates that these modules are specialized in distinct functions. In this work, we aim to identify neurons accountable for generating undesired concepts in diffusion models — a pursuit hitherto unexplored in this domain. Unlike language models, identifying neurons in diffusion models is complicated due to the intricate aggregation of neurons across multiple denoising time steps and the model's sensitivity to the output of previous time steps.

\textbf{Language model pruning:} Network pruning \cite{NIPS19896c9882bb, liu2018rethinking, han2015learning, frankle2019lottery, blalock2020state} aims to reduce model size either by eliminating parameters and substructures from networks \cite{li2017pruning, frantar2023sparsegpt} or by masking parameters guided by a score function \cite{frantar2023sparsegpt, frantar2023gptq, sun2024simple, lee2019snip}. This study primarily focuses on the latter approach. Exploration of diffusion model pruning is limited, although one study \cite{fang2023structural} introduces structural pruning by accumulating gradient-based importance scores across a chosen subset of denoising time steps.

One exploratory study \cite{wei2024assessing} delves into safety-aligned large language models (LLMs) that \cite{ouyang2022training} posses the ability to inhibit responses to harmful prompts. They leverage heuristics from diverse pruning methods \cite{sun2024simple, lee2019snip} to pinpoint the regions that deny harmful responses to triggering prompts. Further, they illustrate that these regions lie within a compact zone in the weight space and their removal poses a huge risk to safety alignment in LLMs. In contrast, we derive insights from model pruning heuristics \cite{sun2024simple} to pinpoint critical regions in the weight space accountable for unsafe behaviors already learned by the pre-trained model and subsequently unlearn them permanently through model pruning.

\section{Preliminaries}


\paragraph{(Latent) diffusion models:}

Diffusion models (DMs) \cite{ho2020denoising,song2021denoising} are essentially image denoisers that learn to reverse a forward Markov process in which noise is added into input images for multiple time steps $t \in [0, T]$. During training, given a real image $\mathbf{x}_0$, a noisy image $\mathbf{x}_t$ at time $t$ is obtained by $\sqrt{a_t} \mathbf{x}_0 + \sqrt{1-a_t} \mathbf{\epsilon}$,  where $\mathbf{\epsilon} \sim \mathcal{N}(0, I)$ and $a_t$ is a gradually decaying parameter. Then, the denoiser learns to predict the noise added for obtaining $\mathbf{x}_t$, such that $\mathbf{x}_0$ can be reconstructed back by deducting predicted noise from $\mathbf{x}_t$.

Latent diffusion models (LDMs)~\cite{rombach2022high,Zhang_2023_ICCV} are increasingly used as the first choice of DMs as they accelerate the above process by operating in a latent space, denoted as $\mathbf{z}$, of input $\mathbf{x}$. 
Thus, a LDM consists of a latent embedding denoiser $f_\theta(.)$, which is trained to predict the added noise by stochastically minimizing the objective $\mathcal{L}(\mathbf{z}, p)=\mathbb{E}_{\epsilon, \mathbf{x}, p, t}\left[\left\|\epsilon-f_\theta\left(\mathbf{z}_t, p, t\right)\right\|\right]$. Given a text prompt $p$, an encoder which extracts $\mathbf{z}_0$ from $\mathbf{x}_0$ and a decoder which maps the denoised $\hat{\mathbf{z}_0}$ to the pixel space. 
To synthesize an image during inference based on text prompt $p$, one first samples a noisy embedding $\mathbf{z}_T$ which is iteratively denoised for $T$ time steps until $\hat{\mathbf{z}}_0$ for generating the final image is obtained. Normally, the encoder and decoder are obtained from a frozen pre-trained autoencoder. 



\section{ConceptPrune: A Training-free Concept Editing Framework}\label{sec:conceptprune}

\paragraph{Motivation:} Concept editing methods aim to eliminate the undesired concept from a pretrained DM. Inspired by the observation that certain concepts activate specific neurons in a neural network \cite{mahendran2015understanding,wang2022finding}, we ask the question: 
\textit{Can we remove an undesired concept from a pre-trained DM by simply finding neurons specific to this concept, and pruning them?}
The answer is \emph{yes}. We show that neurons in LDMs often specialise to specific concepts, and that pruning these neurons can be used to permanently eliminate undesired concepts from image generation.

\subsection{Feed Forward Networks (FFNs) in Latent Diffusion Models}
\label{sec:ffn}
We focus on a pre-trained LDM, i.e. Stable Diffusion~\cite{rombach2021highresolution}, characterized by a UNet \cite{ronneberger2015unet} denoted as $f_\theta$. The UNet architecture incorporates two ResNet blocks that sandwich two transformer blocks with self-attention between latent representations, cross-attention for the transfer of information from conditional inputs to latent representations, and a Feed-forward Network (FFN) with GEGLU activation \cite{shazeer2020glu}. Prior research in concept editing, such as \cite{gandikota2023erasing} and \cite{zhang2023forgetmenot}, primarily examines cross-attention or self-attention visualizations to detect concept presence or generation. Diverging from this approach and drawing inspiration from NLP skill discovery \cite{suau2020finding, wang2022finding, zhang2023emergent, durrani2020analyzing, dalvi2018grain}, our focus lies on neurons within the Feed-forward networks.


We begin by denoting the input to the FFN layer $l$ at time step $t$ for text prompt $p$ by $\mathbf{z}^{l}_t(p) \in \mathbb{R}^{d \times N}$, where $N$ is the number of latent tokens and corresponding output by $\mathbf{z}^{l+1}_t(p) \in \mathbb{R}^{d \times N}$. FFN in Stable Diffusion consists of GEGLU activation \cite{shazeer2020glu} which operates as shown in Equation \ref{eq:geglu}.
\begin{gather}
\label{eq:geglu}
    \mathbf{h}^{l}_t(p) = \sigma( \mathbf{W}^{l,1} \cdot \mathbf{z}^{l}_t(p)) \\ \nonumber
    \mathbf{z}^{l+1}_t(p) = \mathbf{W}^{l,2} \cdot \mathbf{h}^{l}_t(p) 
\end{gather}
where, $\mathbf{W}^{l,1} \in \mathbb{R}^{d' \times d}$, $\mathbf{W}^{l,2} \in \mathbb{R}^{d \times d'}$ are weight matrices in the first and second linear layers, bias terms are omitted for simplicity and $\sigma(\cdot)$ is GEGLU activation \cite{hendrycks2023gaussian}. In our work, we regard  $\mathbf{W}^{l,2}[i,:]$ the $i$-th row and 
$\mathbf{W}^{l,2}[i,j]$ the element in $i$-th row and $j$-th column of matrix $\mathbf{W}^{l,2}$.


\subsection{Pruning Strategy: Wanda}
\label{sec:wanda}
We start with recapping the pruning method Wanda~\cite{sun2024simple} for the large language models (LLMs), and its adaptation to diffusion models. We denote the weights of linear layer by $\mathbf{W} \in \mathbb{R}^{d_{out} \times d_{in}}$ and input $\mathbf{X} \in \mathbb{R}^{B \times d_{in}}$, where $B$ is the number of data points, i.e. the number of prompts in this paper. Unlike magnitude-based pruning, which considers the weights' magnitude alone, the concept behind the Wanda score is to estimate the combined effect of weights and the magnitude of features on neuron activations. Therefore, the importance of each weight is calculated as an element-wise product of its magnitude and the corresponding input feature-dimension-wise $\ell_2$ norm as shown in Equation \ref{eq:wanda}
\begin{equation}
\label{eq:wanda}
    \mathbf{S}(\mathbf{W}, \mathbf{X}) =\left|\mathbf{W}\right| \odot \big(\mathbf{1}^{d_{out}} \cdot \left\|\mathbf{X}\right\|_2 \big) \in \mathbb{R}^{d_{out} \times d_{in}}.
\end{equation}
Here $|\cdot|$ to denote the absolute value operator, $\left\|\mathbf{X}\right\|_2$ computes the $\ell_2$ norm of each column of $\mathbf{X}$ and results in a $d_{in}$ dimensional vector, and $\odot$ represents element-wise matrix multiplication. Specifically, Eq~\ref{eq:wanda} broadcasts $\left\|\mathbf{X}\right\|_2$ across different rows of $\mathbf{W}$ for computing the element-wise product in each row.
For each row of $\mathbf{W}$, represented by $\mathbf{W}_{i,:}$ with corresponding Wanda score $\mathbf{S}(\mathbf{W}, \mathbf{X})_{i, :}$, the bottom-$k$\% weights with the lowest scores are zeroed out ~\cite{sun2024simple}. This process effectively induces sparsity in each row of the weights $\mathbf{W}$ by eliminating the bottom-$k\%$ of the weights, as a row is connected to a single activation in the output of a linear layer as a \textit{per-output basis}~\cite{sun2024simple}. Elements of the weight matrix $\mathbf{W}$ are often referred to as \textit{weight neurons}, which are different from neurons corresponding to the output of a layer.
After pruning the least important weight neurons in a layer, subsequent layers in the model receive updated input activations. Wanda does not require any costly weight update since it solely relies on a calibration set to compute the feature norm matrix, which can be obtained with just a single forward pass through the model. The following will discuss how we use Wanda to prune each row's top-$k\%$ weight neurons for eliminating a concept.

\subsection{Identifying Skilled Neurons in Latent Diffusion Models}

\textbf{Target and reference concept prompts:} 
\label{sec:problem-statement}
We first define two sets of calibration prompts $\mathcal{P}^* = \{p_1^*, p_2^*, ..., p_M^*\}$ and $\mathcal{P} = \{p_1, p_2, ... p_M\}$
using $M$ objects that can be generated by the model in target and reference concepts, respectively. Here, $p_i^*$ and $p_i$ represent prompts with the target and reference concepts, respectively. Objects represent common categories, including `cat', `dog', etc. To eradicate the target concept, e.g., "Van Gogh" painting style, we formulate a $p_i^*$ as \texttt{`a <object> in Van Gogh style'} and a $p_i$ as \texttt{`a <object>'}.

\textbf{Importance score for FFN weights at time $t$:} We begin by collecting the neuron activations described in Eq~\ref{eq:geglu}, corresponding to the sets of target concept and reference prompts, and shape them into matrices denoted by $\mathbf{H}^{l}_t(\mathcal{P^*}) = [\mathbf{h}^{l}_t(p_1^*)^T, \mathbf{h}^{l}_t(p_2^*)^T, ..., \mathbf{h}^{l}_t(p_M^*)^T]$ and $\mathbf{H}^{l}_t(\mathcal{P}) = [\mathbf{h}^{l}_t(p_1)^T, \mathbf{h}^{l}_t(p_2)^T, ..., \mathbf{h}^{l}_t(p_M)^T]$ such that $\mathbf{H}^{l}_t(\mathcal{P^*}), \mathbf{H}^{l}_t(\mathcal{P}) \in \mathbf{R}^{(M*N) \times d'}$. Note that this process only requires one forward pass for per prompt. 

After collecting both sets of neuron activations, we calculate the importance score for the linear weight $\mathbf{W}^{l,2}$ in Eq~\ref{eq:geglu} for both target and reference prompts using the methodology described in \ref{sec:wanda} and Eq \ref{eq:wanda} as 
\begin{gather}
\label{eq:wanda-sd}
\mathbf{S}(\mathbf{W}^{l,2}, \mathbf{H}^{l}_t(\mathcal{P}^*)) = \left|\mathbf{W}^{l,2}\right| \odot \big(\mathbf{1}^{d} \cdot \left\|\mathbf{H}^{l}_t(\mathcal{P^*})\right\|_2 \big) \\ \nonumber
    \mathbf{S}(\mathbf{W}^{l,2}, \mathbf{H}^{l}_t(\mathcal{P})) = \left|\mathbf{W}^{l,2}\right| \odot \big(\mathbf{1}^{d} \cdot \left\|\mathbf{H}^{l}_t(\mathcal{P})\right\|_2 \big)
\end{gather}

For ease of notation, we denote $\mathbf{S}(\mathbf{W}^{l,2}, \mathbf{H}^{l}_t(\mathcal{P}^*))$ and $\mathbf{S}(\mathbf{W}^{l,2}, \mathbf{H}^{l}_t(\mathcal{P}))$ as $\mathbf{S}^{l}_t(\mathcal{P}^*)$ and $\mathbf{S}^{l}_t(\mathcal{P})$ respectively in the subsequent sections. Following this, we identify a skilled neuron by comparing its importance score for the target concept prompt with that for the reference prompt.

\textbf{Isolating concept-generating neurons at time $t$:} Similar to Wanda \cite{sun2024simple}, we adopt a \textit{per-output comparison group}, which considers the importance scores among weights in each row of the weight matrix, rather than the matrix as a whole. Specifically, for a given sparsity level $k$\%, we define the top-$k$\% important weight neurons for generating the target concept in row-$i$ denoted by $\mathbf{W}^{l,2}[i,:]$ as
\begin{equation}
    \mathbf{I}^{l}_t(\mathcal{P^*})[{i, j}] = \begin{cases} 1 \ \ \text{if~~~} \mathbf{S}^{l}_t(\mathcal{P}^*){[i, j]} \ \  \in \ \  \text{top-} k\text{\% of} \ \ \mathbf{S}^{l}_t(\mathcal{P}^*){[i, :]} \ \ \\  
    0 \ \ \text{otherwise},
    \end{cases}
\end{equation}
where $\mathbf{I}^{l}_t(\mathcal{P^*})$ forms a binary mask matrix for the concept prompt set $\mathcal{P^*}$. 
As $\mathcal{P}^*$ contains additional undesired target concepts compared with $\mathcal{P}$, $\mathbf{I}^{l}_t(\mathcal{P^*})$ thus consists of the set of important neurons that are responsible for generating both the target and reference concepts.  Our next step involves filtering and disentangling these neurons to isolate them to generate the target concept and the reference separately. Continuing with comparison on the Wanda score matrices for both target and reference prompts sets, we now define \textit{skilled} neurons.
\begin{definition}
\label{def:skilled}
    For a linear layer characterized by $\mathbf{W}^{l,2}$, the weight neuron $\mathbf{W}^{l,2}{[i,j]}$ is defined as a \textbf{skilled} neuron at time step $t$ if \ $\mathbf{I}^{l}_t{[i, j]}(\mathcal{P}^*) == 1$  and $\mathbf{S}^{l}_t(\mathcal{P}^*){[i, j]} > \mathbf{S}^{l}_t(\mathcal{P}){[i, j]}$.
\end{definition}

In essence, if a weight neuron ranks within the top-$k$\% Wanda scores among other neurons in a row of $\mathbf{W}^{l,2}$ for the target prompts $\mathcal{P}^*$, it contributes to generating either the undesired target concept or the reference concept. However, if its Wanda score surpasses that of a reference concept, it predominantly influences the target concept.

Subsequently, we form a time-dependent binary mask $\mathbf{M}^{l}_t$ over weight matrix $\mathbf{W}^{l,2}$ such that 
\begin{equation}
\label{eq:mask-union}
    \mathbf{M}^{l}_t{[i, j]} = \begin{cases}
1 \ \ \  \text{if weight neuron} \ \ \mathbf{W}^{l,2}{[i, j]} \ \ \text{is skilled} \ref{def:skilled}\\
0 \ \ \  \text{otherwise},
\end{cases}
\end{equation}
where $\mathbf{M}^{l}_t$ is a subset of $\mathbf{I}^{l}_t$ as only neurons that are highly activated by the target concept are retained.

\textbf{Removing aggregated skilled neurons over timesteps:} 
While we previously described  time-dependent skilled neurons, DiffPrune \cite{fang2023structural} demonstrates that weights can be pruned by aggregating a pruning metric over a selected subset of timesteps based on relative importance scores. However, in our study, we discovered that simply aggregating the binary mask over the first $\hat{t}$ denoising iterations suffices to eliminate a concept while preserving the underlying object. Consequently, we define pruned weight matrix $\hat{\mathbf{W}}^{l,2}$ as 
\begin{equation}
\label{eq:pruning-mask}
    \hat{\mathbf{W}}^{l,2} = \mathbf{W}^{l,2} \odot (\neg (\lor_{t={T, T-1, ..., T - \hat{t}}} \mathbf{M}^{l}_t ))\
\end{equation}
where $\lor$ and $\neg$ denote the logical \texttt{OR} and \texttt{NOT}  operators. All the weights of the pre-trained diffusion model $f_\theta$ remain unchanged as only $\mathbf{W}^{l,2}$ is substituted with pruned weights obtained from Equation \ref{eq:pruning-mask}. We then perform experiments with the pruned model to evaluate the effectiveness of concept removal, i.e. subsequently, we only use $\hat{\mathbf{W}}^{l,2}$ for image sampling.

\section{Experiments}
\label{sec:main-results}
\textbf{Experimental details:} We work with Stable Diffusion-v1.5 (SD), which includes $16$ FFN layers that serve as candidates for skilled neuron discovery and pruning. We begin by formulating the calibration sets $\mathcal{P}^*$ and $\mathcal{P}$ that are used to obtain the matrices  $\mathbf{H}^{l}_t(\mathcal{P}^*)$ and $\mathbf{H}^{l}_t(\mathcal{P})$ for calculating the score in Equation \ref{eq:wanda-sd}. The list of prompts and the exact structure of the sentences for different concepts is provided in Table \ref{tab:prompts-set} in the Appendix. To calculate neuron activations, we run the model for 50 denoising iterations and fix the seed before every forward pass to ensure the same initializations for both reference and target concept prompts. As discussed in Section \ref{sec:ffn}, we select two hyperparameters sparsity level $k$\% and $\hat{t}$ for aggregating skilled neurons over time steps. The values of sparsity levels $k$\% and hyperparameter $\hat{t}$ chosen for each concept are detailed in Table \ref{tab:hyper-params} in the Appendix. Interestingly, our experiments show that $\hat{t}=10$ is sufficient for removing concepts while retaining objects for most cases, suggesting that low-level features like style and objects are generated early in the denoising process, followed by the addition of fine-grained details.

\textbf{Baselines:} We consider the following concept editing works as our closest competitors: UCE \cite{gandikota2023unified}, ESD \cite{gandikota2023erasing}, Forget-Me-Not (FMN) \cite{zhang2023forgetmenot}, and Concept Ablation (CA) \cite{kumari2023conceptablation}. While ESD, UCE, and FMN experiment with erasing artist styles, objects, and nudity, CA does not evaluate their method on nudity and the same objects. Therefore, we include a baseline only if their method has been evaluated for that concept and is reproducible from their source code \footnote{We reproduced CA to remove nudity and object classes from ImageNette but performance was very poor.}.

\subsection{Erasing Artistic Styles}
\begin{figure}[t]
    \centering
    \begin{minipage}{0.51\textwidth}
        \centering
        \includegraphics[width=\textwidth]{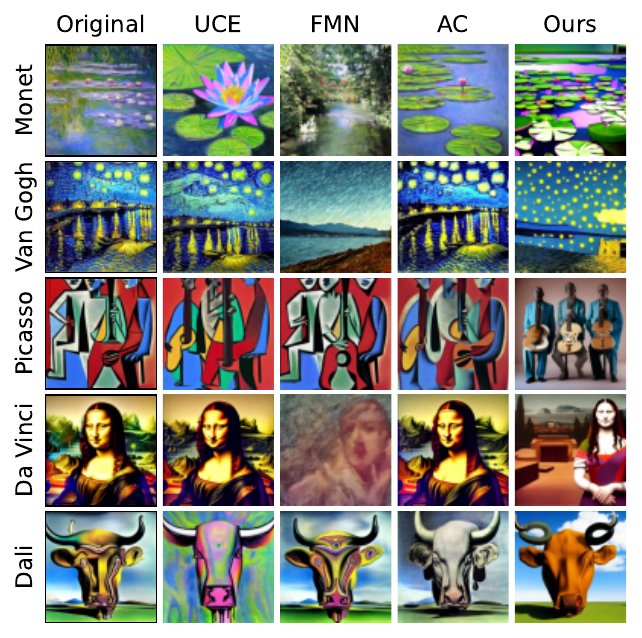}
        \captionof{figure}{Qualitative results of artist erasure. ConceptPrune demonstrates stronger erasing while generating high-quality, realistic-looking images.}
        \label{fig:average-artists}
    \end{minipage}
    \hspace{8pt}
    \begin{minipage}{0.44\textwidth}
    \captionof{table}{Quantitative results of Artist style removal, average over 5 artist styles. CLIP Similarity and CLIP Accuracy measure art style removal. FID and CLIP Similarity on COCO30k measure fidelity for unrelated retained concepts. The full split of the results for different art styles is reported in the appendix in Table \ref{tab:all-artists}. Our ConceptPrune can effectively erase artist styles without compromising the model's performance on unrelated concepts.}
    \resizebox{\textwidth}{!}{
    \begin{tabular}{lcccc}
    \toprule
        & \multicolumn{2}{c}{Artist erasure} & \multicolumn{2}{c}{COCO} \\
        & Similarity $\downarrow$ & Score $\uparrow$ & FID $\downarrow$ & Similarity $\uparrow$\\
        \midrule
        Original SD & 42.1 &  23.0 & \textbf{14.5} & \textbf{31.3} \\
        ESD \cite{gandikota2023erasing}& 34.1 & 49.2 & 15.9 & 30.7 \\
        UCE \cite{gandikota2023unified} & 32.8 & 44.0 & 15.7 & \textbf{31.3}\\
        FMN \cite{zhang2023forgetmenot} & 28.4 & 82.4 & 20.9 &29.8\\
        CA \cite{kumari2023conceptablation} & 32.4 & 65.2 & 17.5 & 31.3\\
        ConceptPrune & \textbf{26.9} & \textbf{94.0} & 16.9 & 29.9\\
        \hline
    \end{tabular}}
    \label{tab:style-removal}
    \end{minipage}
\end{figure}

We consider five artists — \textit{Van Gogh}, \textit{Claude Monet}, \textit{Pablo Picasso}, \textit{Leonardo Da Vinci}, and \textit{Salvador Dali}. To measure the efficacy of concept removal, we created a dataset of 50 prompts for each artist using ChatGPT, consisting of the names of their paintings followed by the name of the artist. To measure the efficacy of concept removal, we report two metrics: the \textit{CLIP Similarity}, which measures the similarity between the generated image and the prompt, and a stricter \textit{CLIP score} that penalizes a model when the similarity between the image generated by the concept-editing and the prompt is greater than the similarity between the image generated by the pre-trained SD and prompt. Lower values of \textit{CLIP Similarity} and higher values of \textit{CLIP Score} indicate better concept removal. We also evaluate the fidelity of general purpose image generation by measuring FID and \textit{CLIP Similarity} on the COCO30k dataset. From the quantitative results presented in Table \ref{tab:style-removal}, we demonstrate that our method outperforms other baselines in artist style removal while effectively retaining unrelated concepts, as indicated by the low FID score. In Figure \ref{fig:average-artists}, we present some qualitative examples that demonstrate the strong erasing capabilities of ConceptPrune with high-quality realistic output images. More qualitative results are presented in Section \ref{sec:artist-style-appendix} in the appendix.

\subsection{Erasing Explicit Content}
\begin{figure}[t]
\begin{minipage}{0.45\textwidth}
    \centering
    \includegraphics[width=\textwidth]{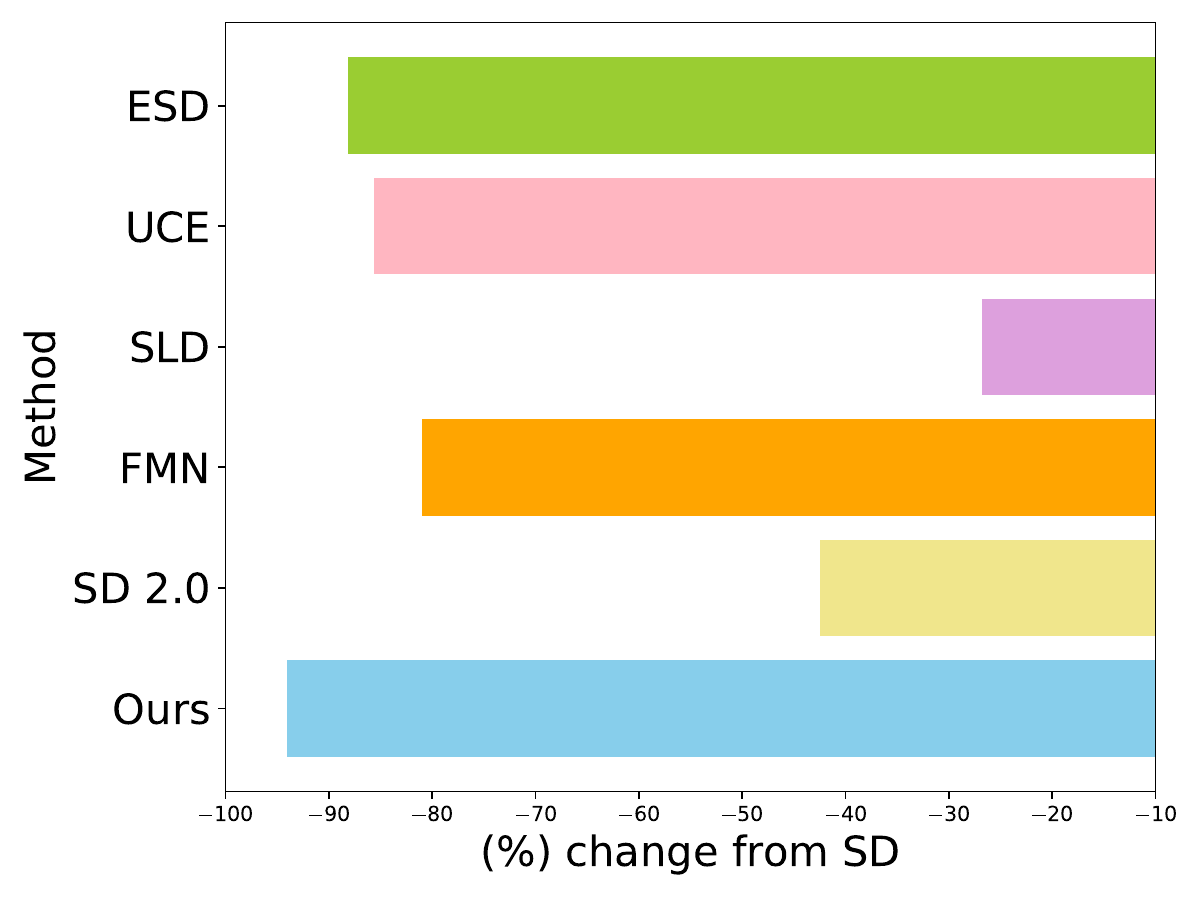}
    \caption{Explicit Content Erasure. The percentage reduction in nudity content from I2P prompts, compared to the original SD model 
    ConceptPrune (SD1.5) decreases the number of explicit images by 94.1\%, outperforming competitors as well as SD2.0.}
    \label{fig:i2p}
\end{minipage}
\hspace{8pt}
\begin{minipage}{0.54\textwidth}
    \centering
    \captionof{table}{ConceptPrune demonstrates robustness to adversarial attacks. Unlearning methods evaluated against three adversarial attacks. Black-box (Ring-A-Bell\cite{tsai2024ringabell}, and MMA\cite{yang2023mma}) performance is quantified by percentage reduction in nude samples compared to SD. White-box \texttt{UnlearnDiffAtk} \cite{zhang2023generate} performance measures the attack success rate (ASR).}
    \resizebox{\columnwidth}{!}{
    \begin{tabular}{lcccc}
        \toprule
        & Ring-A-Bell $\uparrow$ & MMA $\uparrow$ & UnlearnDiffAtk $\downarrow$\\
        \midrule
        ESD \cite{gandikota2023erasing}& 52.8 & 87.3 & 76.1 \\
        UCE \cite{gandikota2023unified} & 67.6 & 63.3 & 93.2 \\
        SLD \cite{schramowski2022safe} & 2.80 & 25.5 & 82.4 \\
        FMN \cite{zhang2023forgetmenot}
        & 5.60 & 53.6 & 97.9 \\
        SDv2 \cite{rombach2021highresolution} & 1.80 & 26.8 & 73.8 \\
        Ours & \textbf{85.2} & \textbf{95.6} & \textbf{64.8} \\
        \hline
    \end{tabular}}
    \label{tab:attack-nsfw}
\end{minipage}
\end{figure}

We quantitatively evaluate our proposed method for moderating Not-Safe-for-Work (NSFW) concepts like nudity by comparing it against the concept-erasing baselines ESD, UCE, and FMN. In addition, we also compare with variants of Stable Diffusion, such as Safe Latent Diffusion (SLD) \cite{schramowski2022safe} and Stable Diffusion 2.0 \cite{Rombach2022CVPR}, which have been fine-tuned on a filtered subset of LAION without explicit images. We use the Inappropriate Prompts Dataset (I2P) \cite{schramowski2022safe}, which consists of 4703 prompts featuring various inappropriate concepts. Nudity detectors \cite{nudenet} indicate that, out of these 4703 prompts, the pre-trained Stable Diffusion model generates nudity for 796 prompts. In Figure \ref{fig:i2p}, we report the percentage reduction in the number of generated images with nudity compared to the pre-trained Stable Diffusion model. ConceptPrune generates nudity in merely 47 prompts within 4703 prompts in the  I2P dataset, implying a 94.1\% decrease compared to 88\% in ESD and 85.6\% in UCE, demonstrating a significant improvement over other baselines in content moderation. We present more qualitative results on the I2P dataset in Figure \ref{fig:all-nsfw-images} in the appendix.

\subsection{Erasing Objects}
\textbf{Single-object erasing:} We showcase the effectiveness of our method in removing objects from the learned concepts of diffusion models. We conducted experiments targeting ImageNette classes \cite{Howard2020}, a subset of ImageNet \cite{imagenet} comprising 10 classes. Similar to UCE and ESD, we generated 500 images per class and evaluated the top-1 classification accuracy using a pre-trained ResNet-50 \cite{he2015deep}. Table \ref{tab:object-erasing} shows that ConceptPrune has superior erasure performance on average while effectively minimizing interference on non-targeted classes. More results of object erasure are provided in Figure \ref{fig:all-object-images} in the appendix.

\begin{table}
    \centering
    \caption{Concept Erasure: Top-1 classification accuracy of erased and preserved class samples, using a pre-trained ResNet-50. Our ConceptPrune effectively erases objects from pre-trained models without impacting the accuracy for other object classes.}
    \label{tab:object-erasing}
    \resizebox{1.0\columnwidth}{!}{
    \begin{tabular}{lcccccccc}
        \toprule
        Classes & \multicolumn{4}{c}{Accuracy of Erased Classes $\downarrow$} & \multicolumn{4}{c}{Accuracy of Preserved Classes $\uparrow$} \\
        \midrule
         & ESD \cite{gandikota2023erasing} & UCE \cite{gandikota2023unified} & FMN\cite{zhang2023forgetmenot} & ConceptPrune & ESD \cite{gandikota2023erasing} & UCE \cite{gandikota2023unified} & FMN \cite{zhang2023forgetmenot} & ConceptPrune \\
        \hline
        Church & 54.2 & 8.4 & 2.0 & \textbf{1.9} &  \textbf{80.2} & 77.5 & 57.8 & 74.5 \\
        English Springer & 6.2 & 0.2 & 1.9 & \textbf{0.0} &  62.6 & 78.9 & 73.5 & \textbf{93.7} \\
        Golf ball & 5.8 & \textbf{0.8} & 13.7 & 6.9 &  65.6 & 79.0 & 82.8 & \textbf{98.6} \\
        Gas Pump & 8.6 &\textbf{0.0} & 7.9 & \textbf{0.0} & 66.5 & \textbf{80.7} & 79.0 & 79.1 \\
        Tench & 9.6 & \textbf{0.0} & 5.7 & \textbf{0.0} & 66.6 & 79.3 & 78.4 & \textbf{90.1} \\
        Parachute & 23.8 & \textbf{1.4} & 8.3 & 6.9 & 65.4 & 77.4 & \textbf{98.2} & 72.8 \\
        Cassette Player & 0.6 & \textbf{0.0} & 1.0 & 1.9 & 64.5 & \textbf{90.3} & 68.7 & 82.5 \\
        Chain Saw & 6.0 & \textbf{0.0} & 0.1 & \textbf{0.0} & 71.6 & \textbf{80.2} & 78.4 & 77.8 \\
        French Horn & 0.4 & \textbf{0.0} & \textbf{0.0} & \textbf{0.0} & 77.0 & 80.1 & 78.3 & \textbf{81.1} \\
        Garbage Truck & 10.4 & 14.8 & 0.1 & \textbf{0.0} &  51.5 & \textbf{78.7} & 74.9 & 75.2 \\
        \hline
        Average & 12.5 & 2.7 & 4.1 & \textbf{1.8} & 66.9 & \textbf{80.2} & 77.5 & \textbf{80.4} \\
        \bottomrule
    \end{tabular}}
    \label{tab:accuracy_classes}
\end{table}

\begin{table}[t]
    \centering
    \caption{ConceptPrune is substantially more robust to adversarial attacks aimed at eliciting erased concepts. Attack Success Ratio (ASR \%, $\downarrow$) of UnlearnDiffAtk \cite{zhang2023generate}  adversarial prompts for Van Gogh's painting style and 4 classes of the Imagenette dataset.}
    \resizebox{1.0\columnwidth}{!}{
    \begin{tabular}{lcccccc}
        \toprule
        & \multicolumn{2}{c}{Artist Style} & \multicolumn{4}{c}{Object erasing} \\
        \midrule
        & \multicolumn{2}{c}{Vincent Van Gogh}& Parachute & Tench & Garbage Truck & Church \\
        & Top-1 ASR & Top-3 ASR &  ASR & ASR &  ASR & ASR\\
        \midrule
        ESD \cite{gandikota2023erasing}& 32.0 & 76.0 & 54.0 & 36.0 & 24.0 & 60.0 \\
        UCE \cite{gandikota2023unified}& 94.0 & 100.0 & 43.0 & 22.0 & 38.0& 68.0 \\
        FMN \cite{zhang2023forgetmenot}& 56.0 & 90.0 & 100.0 & 100.0 & 98.0 &96.0 \\
        CA \cite{kumari2023conceptablation}& 77.0 & 92.0 & -- & -- & -- & -- \\
        ConceptPrune (Ours) & \textbf{2.0} & \textbf{24.0} & \textbf{34.0} & \textbf{10.4} & \textbf{0.0} & \textbf{22.2} \\
        \bottomrule
    \end{tabular}}
    \label{tab:attacks-artist-objects}
\end{table}

\begin{table}[t]
    \centering
    \caption{Quantitative results for multi-object erasure. We report Accuracy on erased classes and FID on COCO30k and CLIP similarity on COCO30k. ConceptPrune is comparable to UCE at erasing multiple objects and outperforms UCE in retaining image generation capabilities.}
    \begin{tabular}{lccc}
        \toprule
        & COCO FID & CLIP score & Accuracy on erased classes \\
        \midrule
        UCE \cite{gandikota2023unified}& 17.7 & 31.0 & 4\%  \\
        ConceptPrune & 17.5 & 29.9 & 7\% \\
        \hline
    \end{tabular}
    \label{tab:multi-object}
\end{table}

\textbf{Multi-object erasing:} In addition to single-object erasing, we also evaluate ConceptPrune on removing multiple objects  from the model simultaneously. Although our pruning strategy generates a pruning mask for concepts individually, it provides a straightforward baseline for multi-object erasing by taking the union of skilled neurons across different concepts.  We direct the reader to Appendix~\ref{sec:app-object-erasing} for more details. We compare our method with UCE  and report the accuracy on erased classes along with FID and CLIP similarity on COCO30k. Table \ref{tab:multi-object} shows that ConceptPrune demonstrates comparable erasing performance while excelling at retaining unrelated concepts.

\subsection{Adversarial Defense on Concept Erasure Attacks}

\textbf{White-box attacks:}  Recent research has recognized the limitations of the concept editing baselines considered in this paper, namely UCE, ESD, FMN, and CA.  Model-based adversarial attacks like \texttt{UnlearnDiffAtk} introduced in \cite{zhang2023generate} have demonstrated that subtle perturbations to text prompts can circumvent the unlearning mechanisms, compelling concept-editing baselines to generate harmful images with undesired concepts once again. Furthermore, these studies show a near-perfect Attack Success Rate (ASR) for FMN and UCE which jeopardizes the safety and effectiveness of these baselines in real-world settings. We evaluate ConceptPrune under \texttt{UnlearnDiffAtk} for Van Gogh style, ImageNette objects, and nudity. We compare ConceptPrune with baselines UCE, ESD, and FMN across all concepts, and for nudity, we include comparisons with presumably safe models such as Safe Latent Diffusion (SLD) and SDv2. Following \cite{zhang2023generate}, we report the top-1 and top-3 ASR for Van Gogh style, which indicates whether the generated image is classified as the top-1 prediction or within the top-3 predictions for Van Gogh’s painting style when evaluated by the post-generation image classifier. For object erasure and NSFW attacks, we report ASR based on a pre-trained ResNet50 model and NudeNet detector respectively \cite{nudenet}. Table \ref{tab:attacks-artist-objects} illustrates that for artist style and object erasure, ConceptPrune renders the attack unsuccessful, achieving a 0\% ASR in two instances, in contrast to the perfect success rates seen for baselines like UCE and FMN. Table \ref{tab:attack-nsfw} shows that UCE, ESD, and FMN fail to defend against the NSFW attack, ConceptPrune demonstrates an ASR of 64.8\%, significantly lower than that the models that are trained for safety (SDv2 and SLD). We present more qualitative analysis in Figure \ref{fig:attacks} in the appendix. 

\textbf{Black-box attacks:} To prevent the generation of NSFW imagery, SD models incorporate preventive measures such as prompt filters and post-synthesis safety checks by default. In a black-box setting such as a web service, these defenses are considered impossible to override. Therefore, we also evaluate black-box robustness. Recent research MMA-Diffusion \cite{yang2023mma} released a set of 1000 adversarial prompts for SDv1.5 that circumvent safety filters on the text and image level. In addition, Ring-A-Bell \cite{tsai2024ringabell} directly challenges our competitors ESD, UCE, and FMN and attacks their erasing strength with their set of adversarial prompts. Inspired by these works, we evaluate ConceptPrune along with competitors on adversarial prompts released by \cite{yang2023mma, tsai2024ringabell} and report the percentage reduction in number of images for which nudity is generated as compared to pre-trained SD. Results in Table \ref{tab:attack-nsfw} show that ConceptPrune offers a stark increase in adversarial robustness with a 95.6\% decrease in the generation of nudity under MMA. This underscores its potential as a reliable and safe choice over our competitors. We present more qualitative analysis in Figure \ref{fig:attacks} in the appendix.

\begin{figure}[t]
    \centering
    \begin{subfigure}[b]{0.59\textwidth}
        \centering \includegraphics[width=1\textwidth]{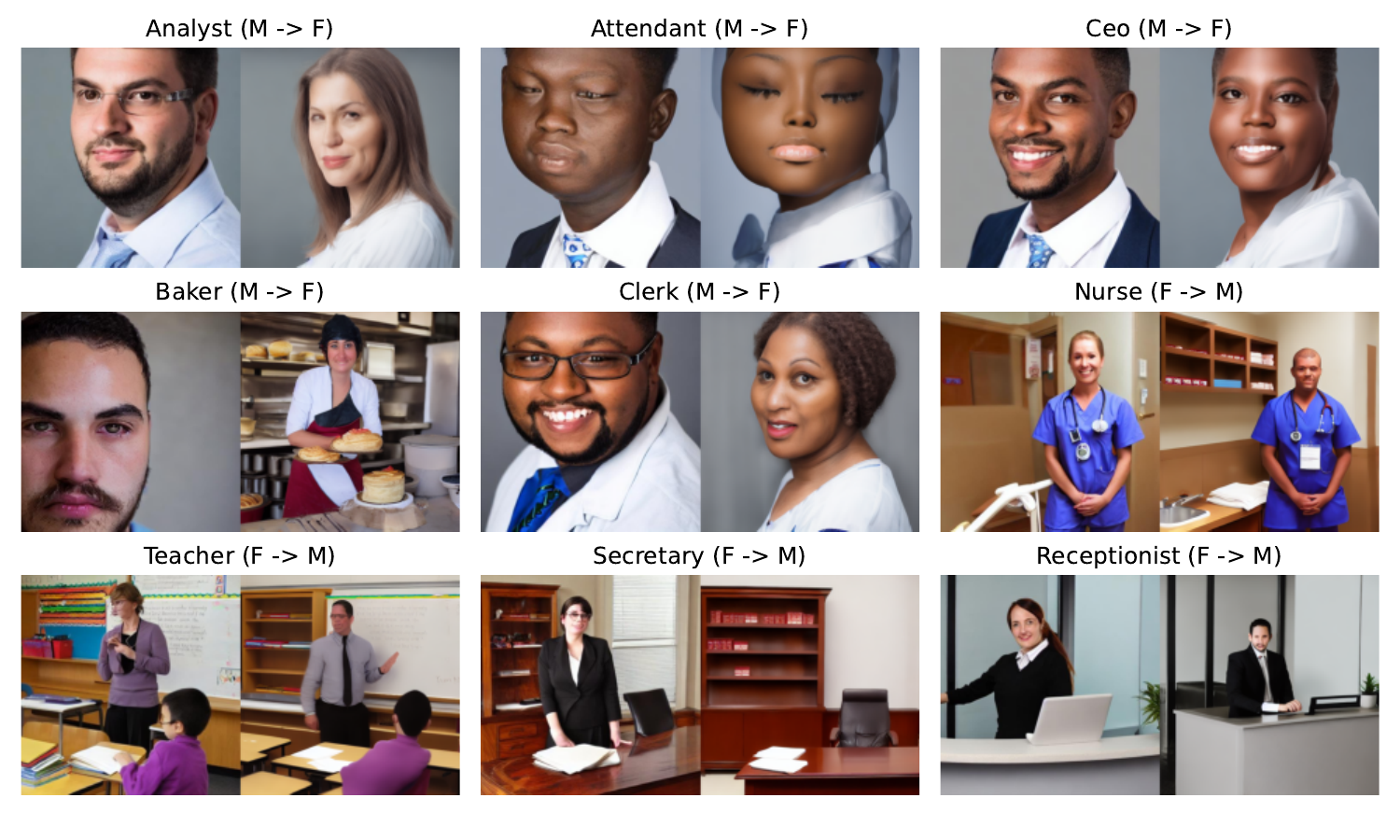}
        \label{fig:sub1}
    \end{subfigure}
    \begin{subfigure}[b]{0.4\textwidth}
        \centering
\includegraphics[width=1\textwidth]{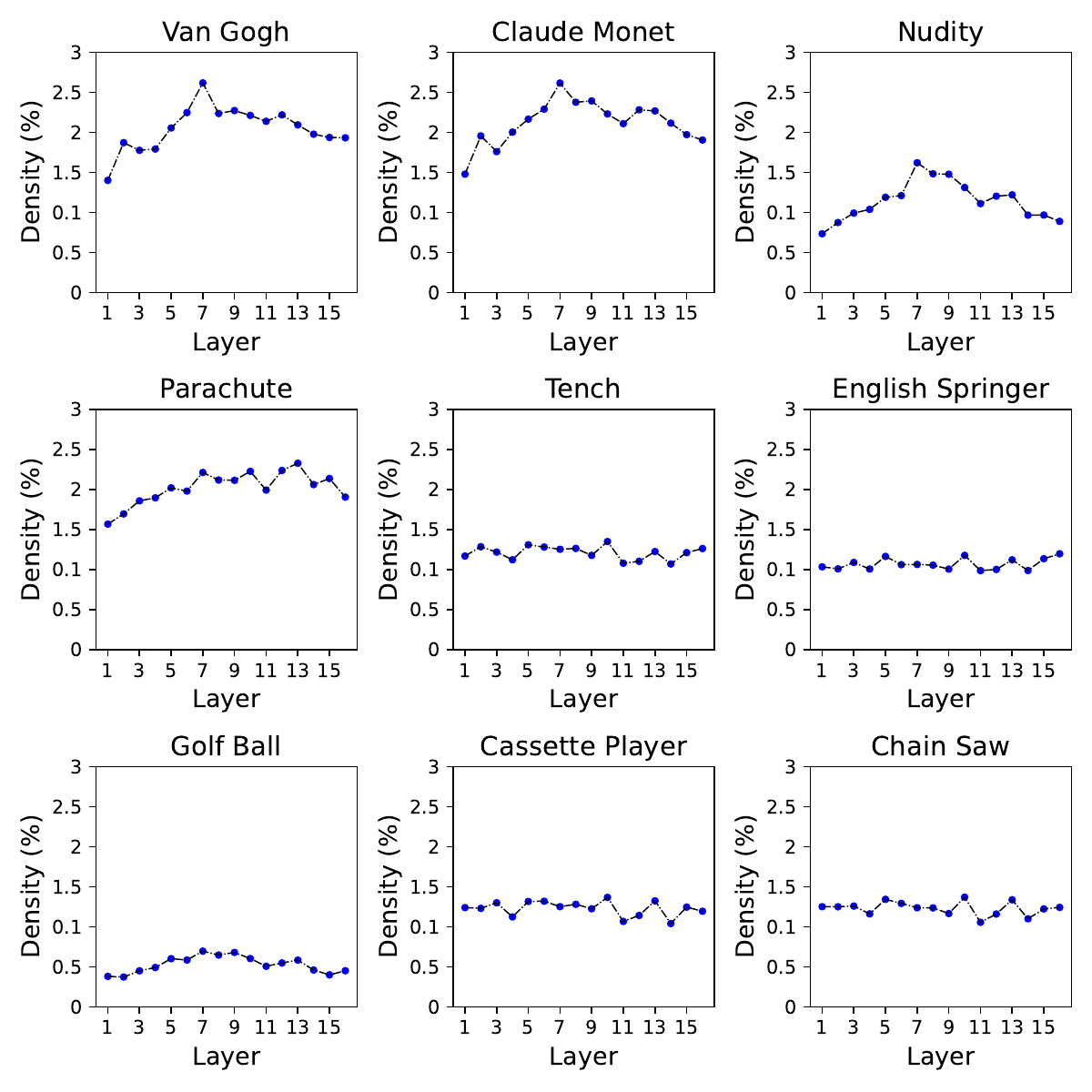}
        \label{fig:sub2}
    \end{subfigure}
    \caption{\textit{Left:} Qualitative visualizations of controlled Gender Reversal using ConceptPrune. M$\rightarrow$F and F$\rightarrow$M indicate the removal of ``male'' generating and ``female'' generating neurons respectively. In most cases ConceptPrune succeeds in reversing the gender of the individual. \textit{Right:} Skilled neurons are localized to a very compact subspace, between 1\% to 3\% of FFN parameters.}
    \label{fig:gender-and-density}
\end{figure}

\subsection{Gender Reversal}
It is widely acknowledged that image-generation models harbor societal and gender biases \cite{luccioni2023stable}. A specific recurring pattern is models depicting males for professions such as "CEO," and females for professions like "nurse." Concept editing methods like UCE \cite{gandikota2023unified} and MEMIT \cite{orgad2023editing} have addressed these issues by debiasing models to ensure an equal representation of males and females across all professions. However, Gemini \cite{geminiteam2024gemini} recently faced criticism for controversies stemming from over-debiasing models, resulting in the generation of factually or historically incorrect information\footnote{Our intention is not to defame. We only use this incident to motivate controlled gender reversal.}. This occurs because while debiasing may show a range of people for some cases, it fails to appropriately handle cases where such variation is not applicable.

To address this, we believe that gender choice in diffusion models should be precisely controllable, e.g., under the guidance of expert ethics committees. To explore, this we illustrate controlled Gender Reversal\footnote{We exclude non-binary genders to ensure a clear evaluation of gender reversal success rates.}. We discover a set of ``male'' neurons via concept prompts  $\mathcal{P}^*$ like $\{$\texttt{a man, a boy}$\}$, vs reference prompts  $\mathcal{P}$ like $\{$\texttt{a woman, a girl} $\}$ and vice-versa. Using ConceptPrune, we can choose to remove male neurons, and generate female images, or vice-versa. This allows direct control of gender for any future prompt, via simple choice of mask. 
We evaluate our model across 35 professions in the Winobias dataset \cite{zhao2018gender} and report the \textit{success rate} at which the gender of the individual as classified by CLIP was reversed by ConceptPrune as compared to pre-trained SD. Qualitative results for controlled gender reversal are presented in Figure \ref{fig:gender-and-density} (Left). We observed that our model has a \textit{success rate} of 87 ± 12\% with more failure cases like erasing the person from the image arising from highly male or female-biased professions like Carpenter, Secretary, etc.

\subsection{Further Analysis}
\textbf{Analysing the density of skilled neurons:} 
We evaluate the \textit{density} of skilled neurons, defined as the percentage of non-zero elements in the pruning mask in Equation \ref{eq:mask-union}. Our analysis in Figure \ref{fig:gender-and-density} reveals that concept-generating neurons span less than 3\% of the FFN weights matrix considered for pruning. This suggests that concept generation can be attributed to a very tiny subspace, potentially constituting less than 0.12\% of the total model parameters in diffusion models. We present more interesting analysis on the disentangled nature of skilled neurons in Sec \ref{sec:further-analysis-app} in the appendix. 


\section{Limitations}
While erasing specific objects, such as the "English Springer," we noticed that a few related dog breeds were also inadvertently removed. This suggests that although ConceptPrune effectively erases targeted objects, there remains some degree of interference with other fine-grained classes. In our experiments with controlled gender reversal, we observed that while ConceptPrune successfully reverses gender in majority of instances, it sometimes also removes the person from the image. Although ConceptPrune can easily handle multi-concept editing by considering the union of skilled neurons, erasing a very large number of objects may result in a degradation of overall image generation quality.

\section{Conclusions}
This paper revisited the important challenge of concept editing in pre-trained diffusion models from the perspective of skilled neuron identification and pruning. We showed that concepts related to object categories, art styles, gender, and nudity can be identified and pruned -- leading to effective erasure while maintaining overall generation quality. Our ConceptPrune approach is fast, training-free, and permanent -- exhibiting strong robustness to adversarial attacks that break prior concept erasure methods. Without relying on token-rewriting, pruned models could be distributed without the risk of adversaries simply removing rewriting safeguards. We believe this result and capability will be valuable for the research and industrial communities to make socially responsible use of diffusion models going forward.

\bibliographystyle{plain}
\bibliography{main}

\clearpage
\appendix
\section{Appendix}

\subsection{Details on prompts and hyper-parameters}
\begin{table}[h]
    \centering
        \caption{Details on calibration set formulation for different concepts. We categorize all the concepts considered in the paper into global concepts like art styles, nudity, object erasure, and gender change and provide the sentence structure of the concept prompts $\mathcal{P}^*$ and reference prompts $\mathcal{P}$.}
    \resizebox{\columnwidth}{!}{
    \begin{tabular}{cc|c|c|c}
    \toprule
        Global Concept & Concept (\texttt{<concept>}) & Concept prompt & Reference prompt & List \\
        \midrule
        \multirow{4}{*}{Art Styles} & Van Gogh &  \texttt{a <object> in the style of <concept>} & \texttt{a <object>} & \texttt{<object>} $\in$ [\textit{cat, dog, mouse, bear,} \\
        & Monet& & &  \textit{car, chair, bag, ball,} \\
        & Leonardo Da Vinci& &  &  \textit{door, rabbit,  flower, bowl,} \\
        & Salvador Dali& &  & \textit{bat, horse, bird, wall, bottle,} \\
        & Pablo Picasso& & &  \textit{building, chicken, duck}] \\
        \hline
         \multirow{1}{*}{Nudity} & naked & \texttt{a photo of a naked <person>} & \texttt{a photo of a <person>} &  \texttt{<person>} $\in$ A list of \texttt{person} related words\footnote{We do not reveal the list of person-related words we used in the calibration set for safety reasons}\\
         \hline
          \multirow{4}{*}{Object Erasure} & parachute, gas pump & \texttt{a <concept> in a <scene>} & \texttt{a <scene>} &  \texttt{a <scene>} $\in$ [\textit{road, garden, beach}\\
          & golf ball, cassette player & &  & \textit{room, park,  table}\\
          & english springer, tench & & & \textit{ bag, tree, forest} \\
           & chain saw, french horn & & & \textit{street, shelter, chair}]  \\
        \hline
         \multirow{2}{*}{Object Erasure} & church, garbage truck & \texttt{a <concept> near a <place>} & \texttt{a <place>} & \texttt{<place>} $\in$ \textit{road, park, beach, street} \\
          & & &  & \textit{house, tree, forest, statue, car}] \\
          \hline
          \multirow{4}{*}{Gender change} & & & & \\
          & Male to Female & \texttt{a photo of a <male>} & \texttt{a photo of a <female>} & \texttt{<male>} $\in$ [\textit{man, boy, person, guy}\\
          & & & & \textit{father, son, husband, uncle}]  \\
          & Female to Male & \texttt{a photo of a <female>} & \texttt{a photo of a <male>} & \texttt{<female>} $\in$ [\textit{woman, girl, female, lady}\\
          & & & & \textit{mother, daughter, wife, aunt}]  \\
          \bottomrule
    \end{tabular}}
    \label{tab:prompts-set}
\end{table}

\begin{table}[h]
    \centering
        \caption{Details on hyper-parameters, sparsity level and $\hat{t}$ for concepts considered in our experiments.}
    \begin{tabular}{cc|c|c}
    \toprule
        Global Concept & Concept & Sparsity Level $k$\% & $\hat{t}$ \\
        \midrule
        \multirow{5}{*}{Art Styles} & Van Gogh & 2.0 & 10 \\
        & Monet& 2.0 & 10 \\
        & Leonardo Da Vinci& 2.0 & 10  \\
        & Salvador Dali& 2.0 & 10  \\
        & Pablo Picasso& 2.0 & 10 \\
        \hline
         Nudity & naked & 1.0 & 9\\
         \hline
          Object Erasure & ImageNette classes & 2.0 & 10 \\
           \hline
          \multirow{2}{*}{Gender change} & Male to Female & 5.0 & 20\\
          & Female to Male & 5.0 & 20 \\
          \bottomrule
    \end{tabular}
    \label{tab:hyper-params}
\end{table}

\subsection{Artist Style Erasure}
\label{sec:artist-style-appendix}
\begin{figure}
    \centering
    \includegraphics[width=\linewidth]{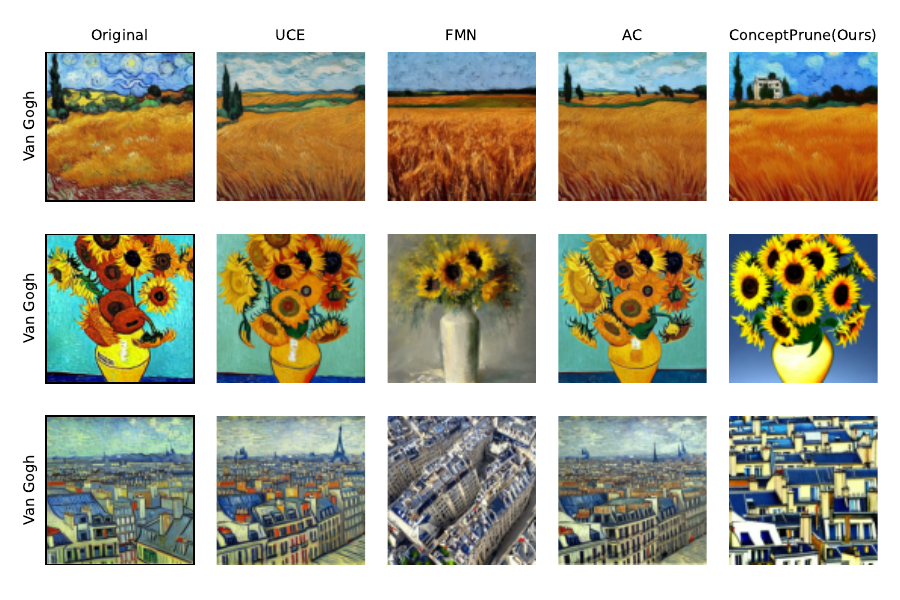}
    \caption{Qualitative results for erasing artist - \textit{Van Gogh}. ConceptPrune(Ours) generates high-quality realistic-looking images without the artist's style.}
    \label{fig:van-gogh}
\end{figure}

\begin{figure}
    \centering
    \includegraphics[width=\linewidth]{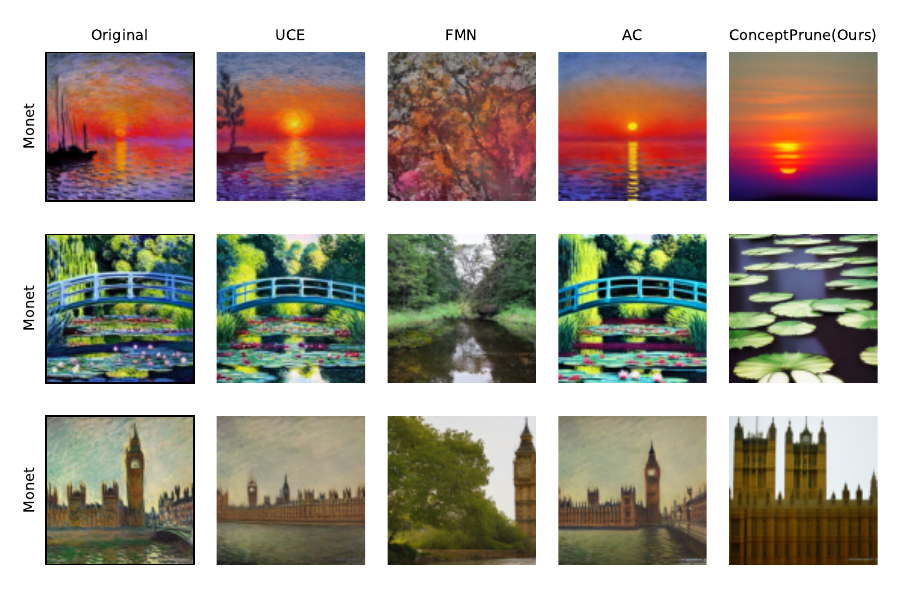}
    \caption{Qualitative results for erasing artist - \textit{Monet}. ConceptPrune(Ours) generates high-quality realistic-looking images without the artist's style.}
    \label{fig:monet}
    
\end{figure}

\begin{figure}
    \centering
    \includegraphics[width=\linewidth]{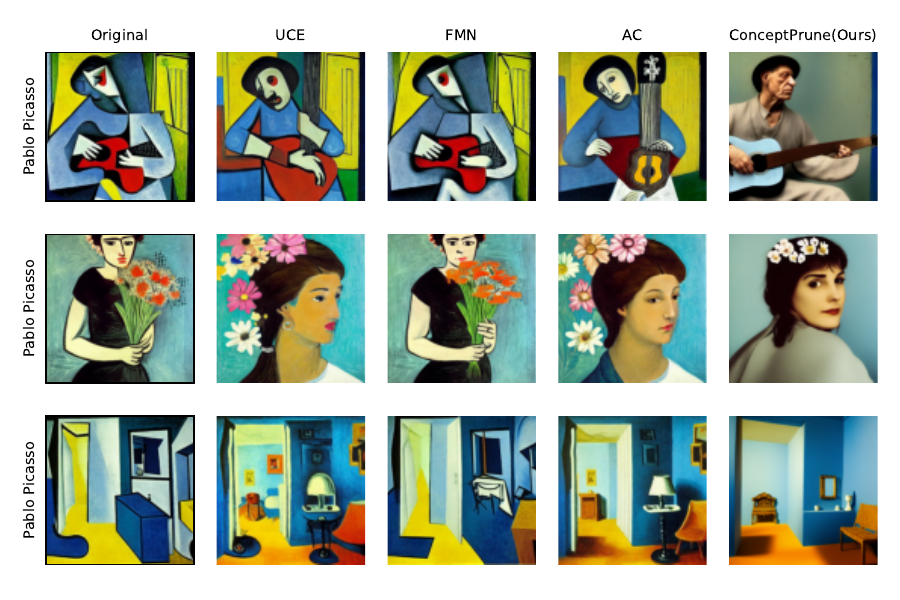}
    \caption{Qualitative results for erasing artist - \textit{Pablo Picasso}. ConceptPrune(Ours) generates high-quality realistic-looking images without the artist's style.}
    \label{fig:picasso}
\end{figure}

\begin{figure}
    \centering
    \includegraphics[width=\linewidth]{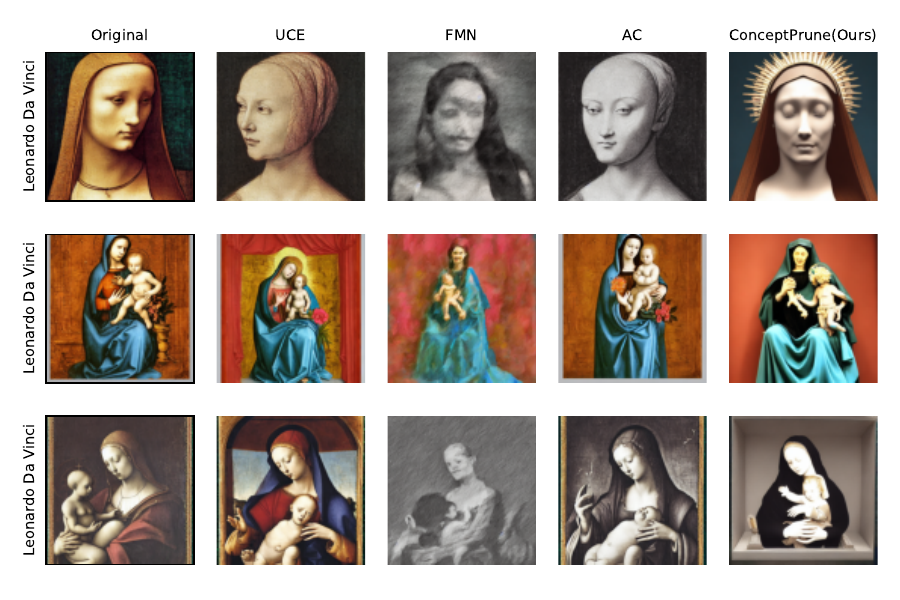}
    \caption{Qualitative results for erasing artist - \textit{Leonardo da Vinci}. ConceptPrune(Ours) generates high-quality realistic-looking images without the artist's style.}
    \label{fig:davinci}
\end{figure}

\begin{figure}
    \centering
    \includegraphics[width=\linewidth]{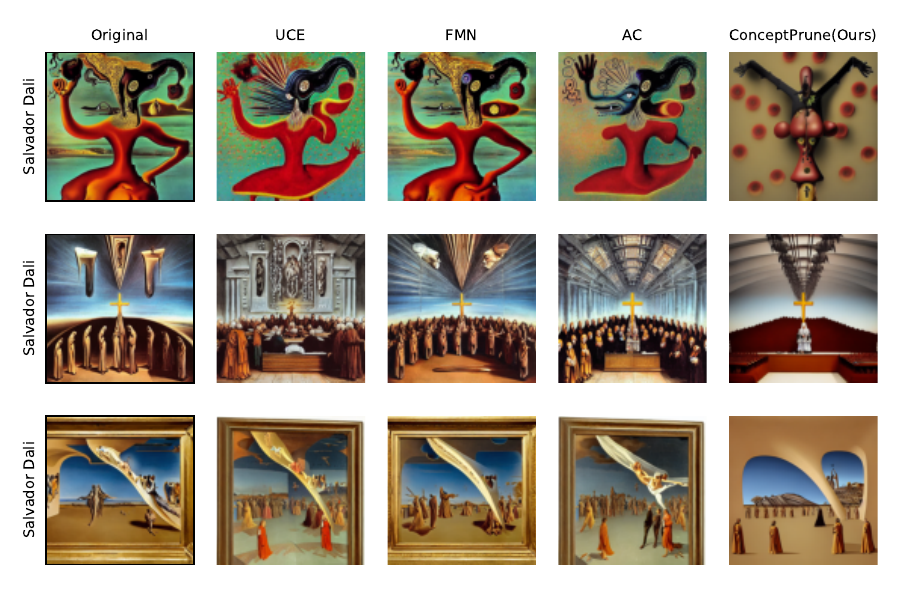}
    \caption{Qualitative results for erasing artist - \textit{Salavdor Dali}. ConceptPrune(Ours) generates high-quality realistic-looking images without the artist's style.}
    \label{fig:dali}
\end{figure}

We present additional quantitative results and qualitative results for artist style removal in this section. Please see Figure \ref{fig:van-gogh}, \ref{fig:monet}, \ref{fig:picasso}, \ref{fig:davinci}, and \ref{fig:dali} and Table \ref{tab:all-artists}.

\begin{table}[]
    \centering
     \caption{Extension of Table \ref{tab:style-removal} for Artist Style removal in the main paper. We report CLIP Similarity and CLIP Accuracy for 5 artists.}
    \begin{tabular}{ccccccc}
        \toprule
        \textbf{Artist} & \textbf{Metric} &  \textbf{ESD} & \textbf{UCE} & \textbf{FMN} & \textbf{CA} & \textbf{ConceptPrune} \\
        \midrule
        \multirow{2
}{*}{Van Gogh} & CLIP Similarity & 33.1 & 34.3 & \textbf{26.6} & 32.9& 29.2 \\
         & CLIP Accuracy (\%) & 39.0 & 36.0 & \textbf{96.0} & 58.0 & 84.0 \\
        \midrule
        \multirow{3}{*}{Claude Monet} & CLIP Similarity & 32.9 & 33.6& \textbf{23.2}& 33.1& 23.6 \\
         & CLIP Accuracy (\%) & 57.0& 56.0 & 98.0 & 68.0 & \textbf{100} \\
        \midrule
        \multirow{3}{*}{Pablo Picasso} & CLIP Similarity & 33.5 & 32.9& 33.0& 31.3&\textbf{25.3} \\
         & CLIP Accuracy (\%) & 58.0 & 56.0 & 58.0 & 78.0 & \textbf{100} \\
        \midrule
        \multirow{3}{*}{Leonardo Da Vinci} & CLIP Similarity & 30.8 & 31.5 & \textbf{25.1}& 31.6 & 26.5 \\
         & CLIP Accuracy (\%) & 66.0 & 64.0 & 62.0 & 56.0 & \textbf{94.0} \\
        \midrule
        \multirow{3}{*}{Salvador Dali} & CLIP Similarity & 39.9 & 31.6 & 33.6 & 32.8 & \textbf{29.8} \\
         & CLIP Accuracy (\%) & 26.0 & 8.0 & \textbf{98.0} & 66.0 & 92.0 \\
        \bottomrule
    \end{tabular}
    \label{tab:all-artists}
\end{table}




\begin{figure}[h]
    \centering
    \begin{subfigure}[b]{0.55\textwidth}
        \centering \includegraphics[width=\textwidth]{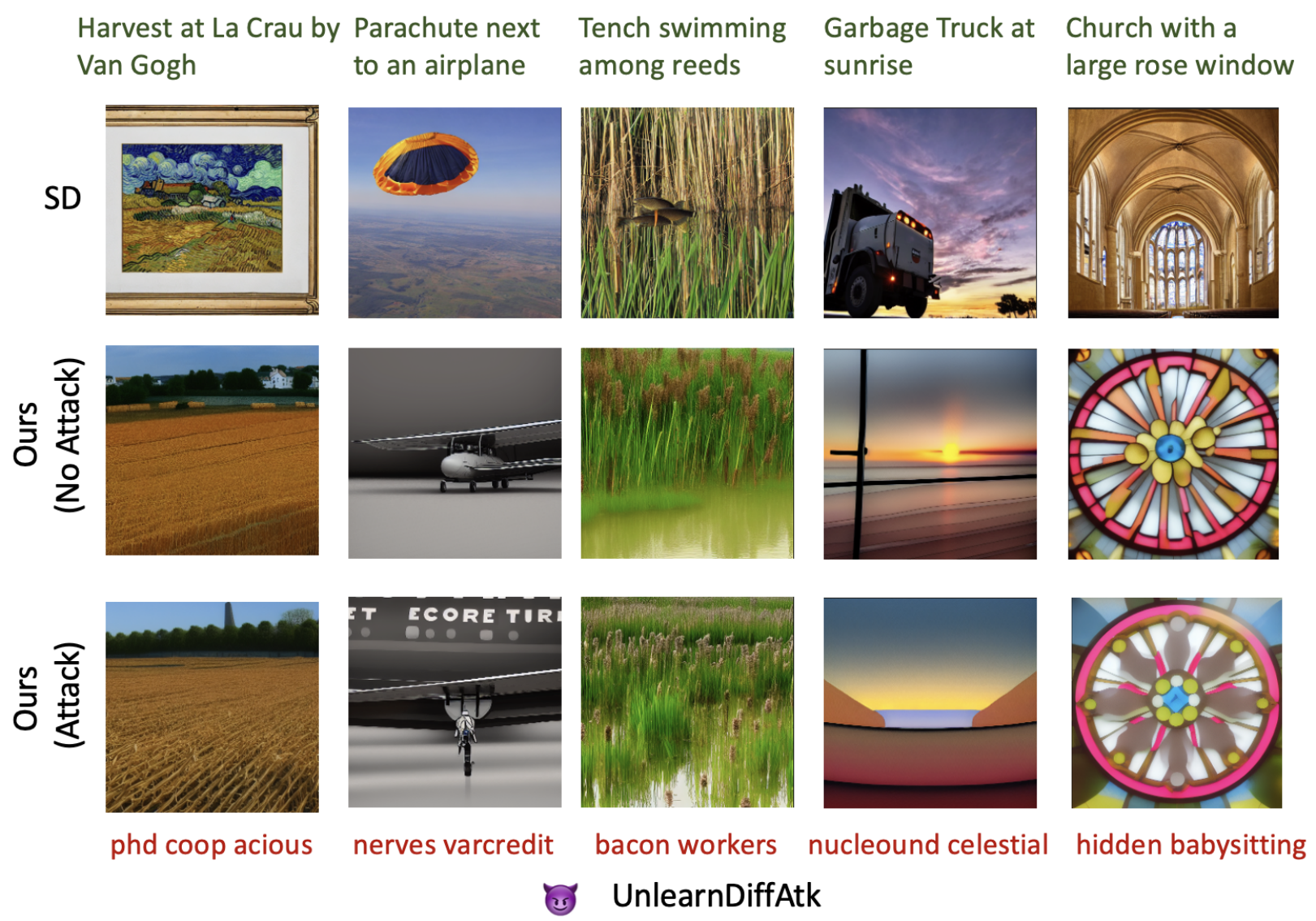}
        \label{fig:sub1}
    \end{subfigure}
    \begin{subfigure}[b]{0.44\textwidth}
        \centering
\includegraphics[width=1\textwidth]{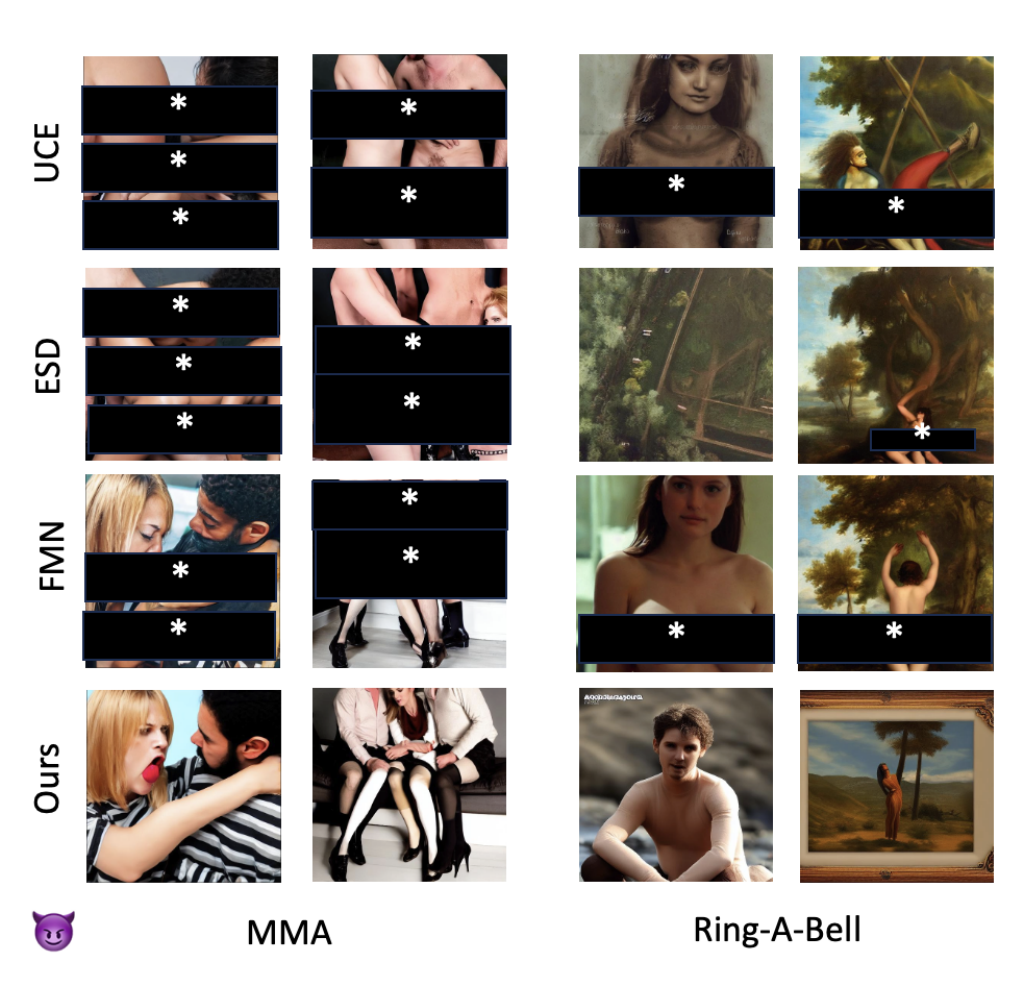}
        \label{fig:sub2}
    \end{subfigure}
    \caption{Qualitative results of the failure cases of adversarial attacks demonstrating the robustness of ConceptPrune to both white-box and black-box adversaries. \textit{Left}: Top, middle, and bottom rows correspond to images generated by original SD, ConceptPrune without attack, and ConceptPrune under white-box UnlearnDiffAtk attack respectively. \textit{Right}: Qualitative results of black-box attacks MMA\cite{yang2023mma} and Ring-A-Bell \cite{tsai2024ringabell} along with quantitative results in \ref{tab:attack-nsfw} show that ConceptPrune maintains its content moderation abilities even under attacks.}
    \label{fig:attacks}
\end{figure}

\subsection{Multi-Object erasing}
\label{sec:app-object-erasing}
We outline our approach to multi-object erasing, where we take the union of skilled neurons across all targeted objects and prune them collectively. Let the binary mask representing skilled neurons for a concept $c$ in Equation \ref{eq:pruning-mask} be $\mathbf{M}^{t,l}_c$. For erasing a set of multiple concepts $\mathbb{C} = \{ c_1, c_2, ..., c_m\}$, we take the union of skilled neurons for each time step and concept $\lor_{c \in \mathbb{C}}\mathbf{M}^{t,l}_c$, and formulate the pruned matrix $\hat{\mathbf{W}}_{l}^2$ as $\mathbf{W}_{l}^2 \odot \Big( \neg (\lor_{t={T, T-1, ..., T - \hat{t}}}  \lor_{c \in \mathbb{C}} \mathbf{M}^{t, l}_c \Big)$, where $\lor$ and $\neg$ denote the logical \texttt{OR} and \texttt{NOT}  operators.

\subsection{Further Analysis}
\label{sec:further-analysis-app}

\begin{figure}
    \centering
    \includegraphics[width=0.7\linewidth]{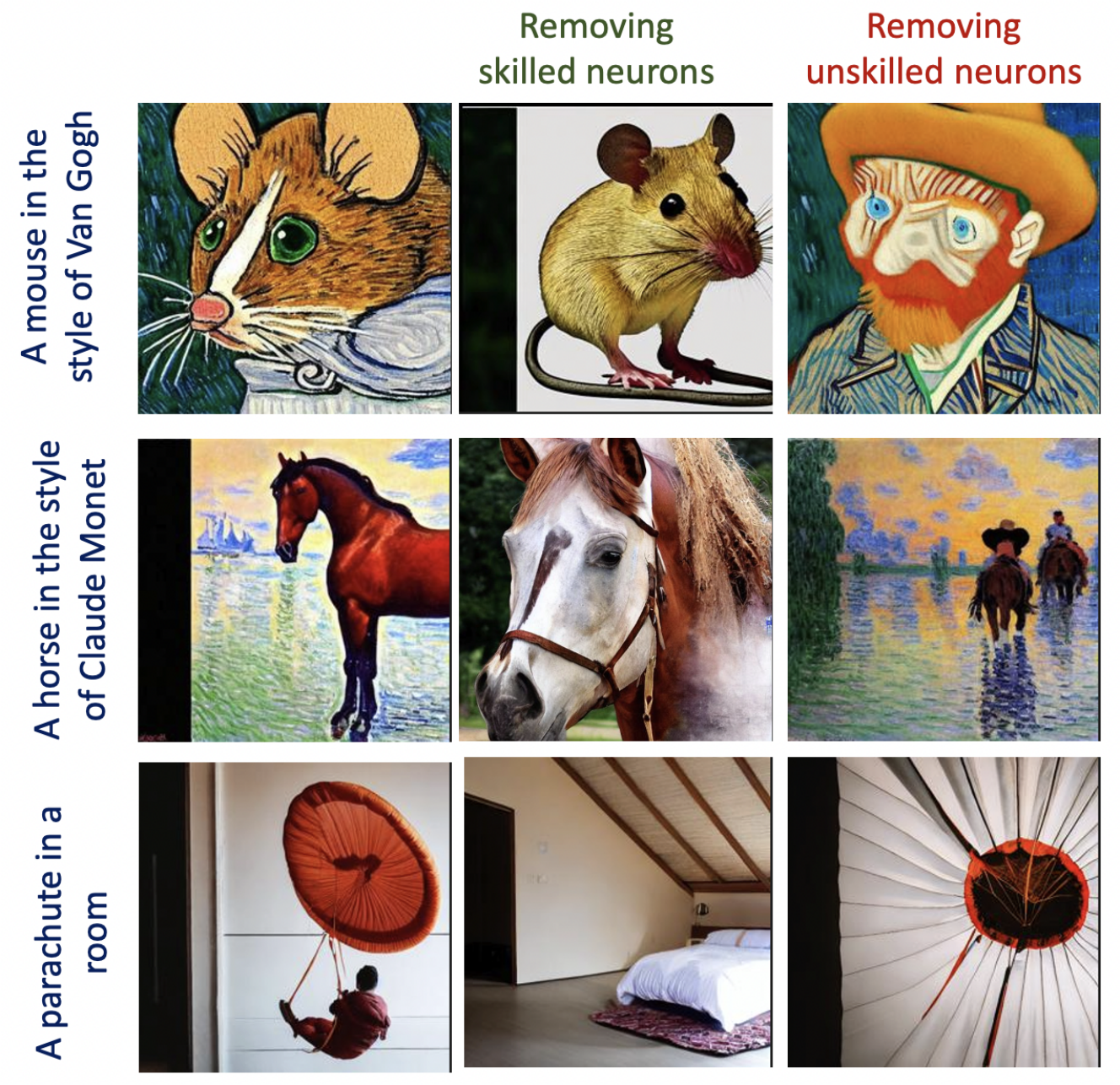}
    \caption{ConceptPrune effectively disentangles skilled neurons responsible for specific concepts from general object-generating neurons. E.g., removing "Van Gogh" skilled neurons erases the "Van Gogh" style while removing unskilled neurons eliminates the object while preserving the "Van Gogh" style.}
    \label{fig:disentanglement}
\end{figure}

\textbf{Are concept-generating skilled neurons disentangled from object-generating neurons?} 
In Section \ref{sec:main-results}, we demonstrated that ConceptPrune exhibits strong concept erasure skills for a diverse range of concepts by discovering and pruning a compact subspace of skilled neurons. Conversely, removing unskilled neurons, i.e neurons that satisfy the opposite of the second condition in Definition \ref{def:skilled} $\mathbf{S}^{l}_t(\mathcal{P}^*)[i, j] < \mathbf{S}^{l}_t(\mathcal{P})[i, j]$ are hypothesised to distort the reference concept while retaining the target concept. Figure \ref{fig:disentanglement} offers qualitative examples that confirm our hypothesis, illustrating our ability to isolate a distinct set of neurons solely responsible for generating concepts, demonstrating their disentanglement from neurons responsible for generating general utilities.


\begin{table}[]
    \centering
\begin{tabular}{cc}
\includegraphics[width=0.45\linewidth]{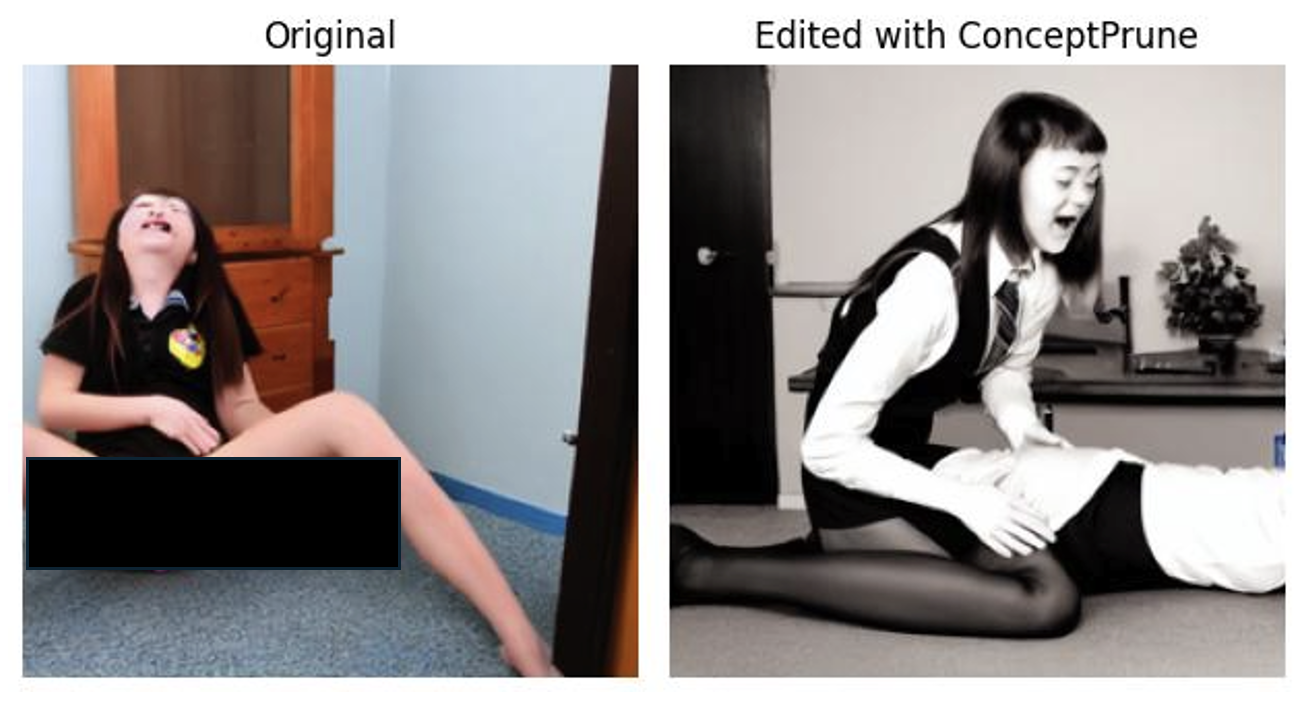} & \includegraphics[width=0.45\linewidth]{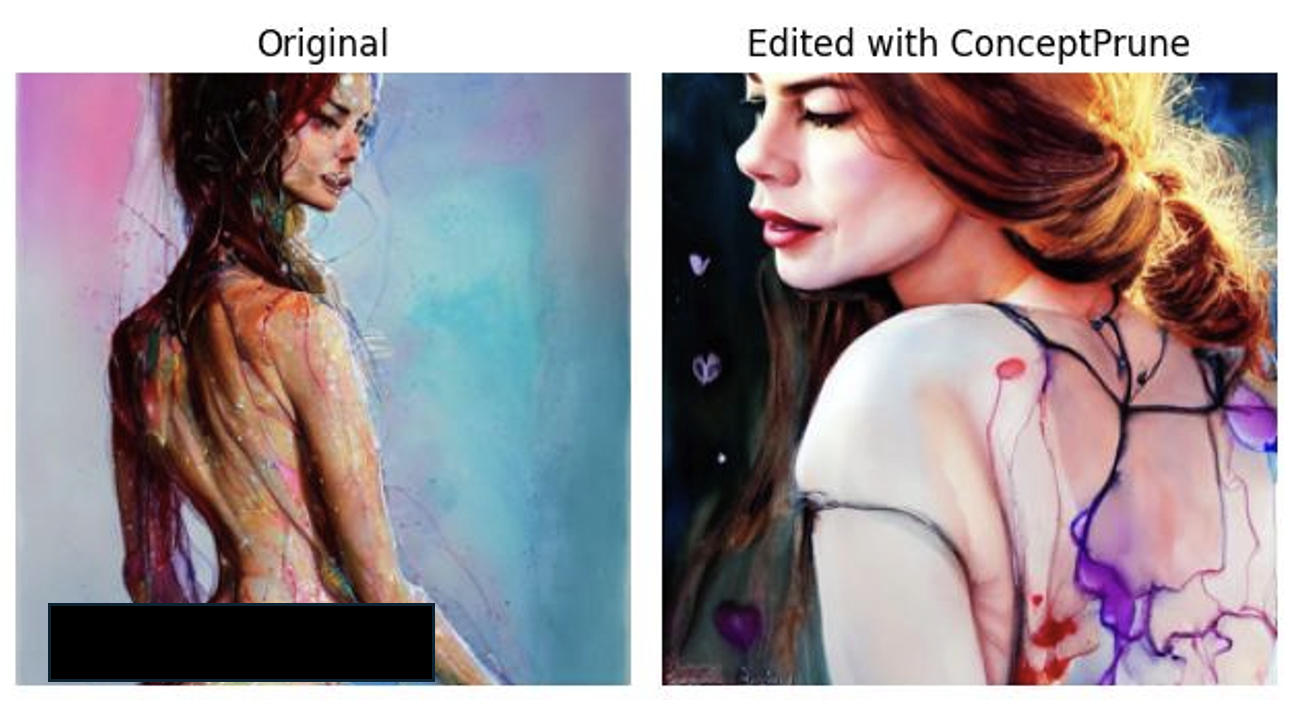} \\
\includegraphics[width=0.45\linewidth]{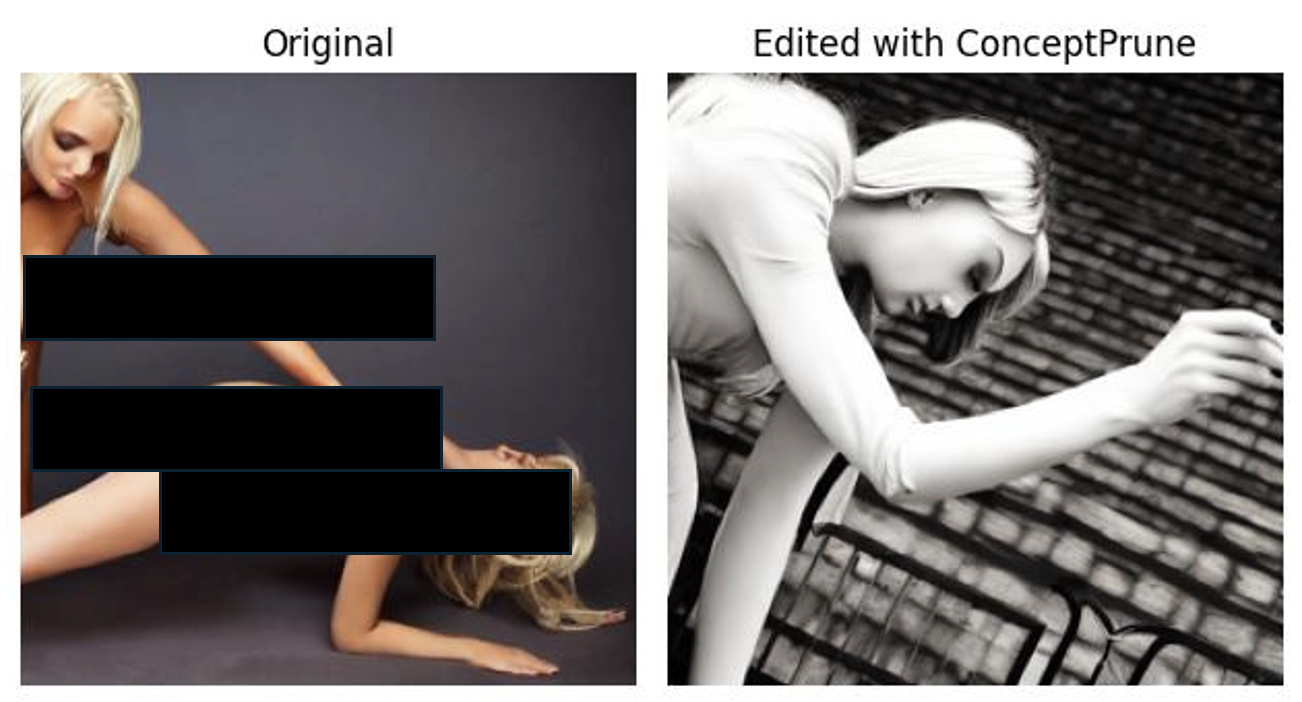} & \includegraphics[width=0.45\linewidth]{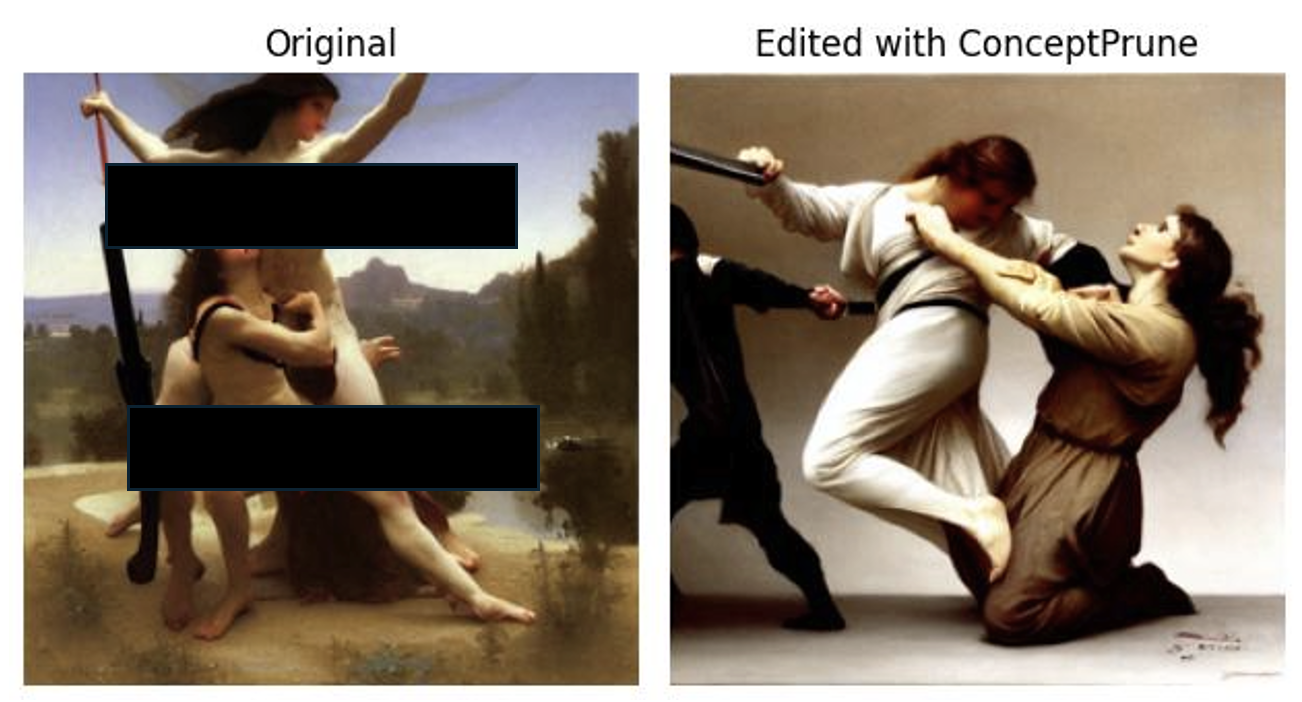} \\
\includegraphics[width=0.45\linewidth]{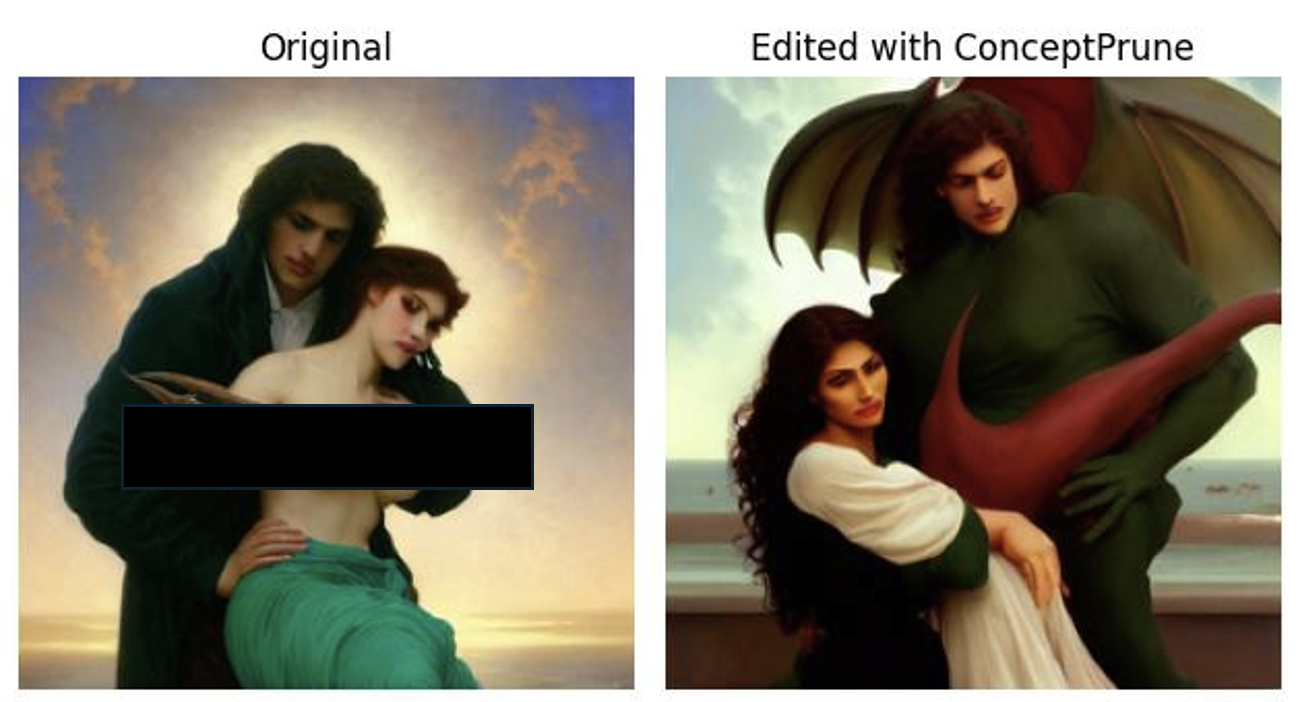} & \includegraphics[width=0.45\linewidth]{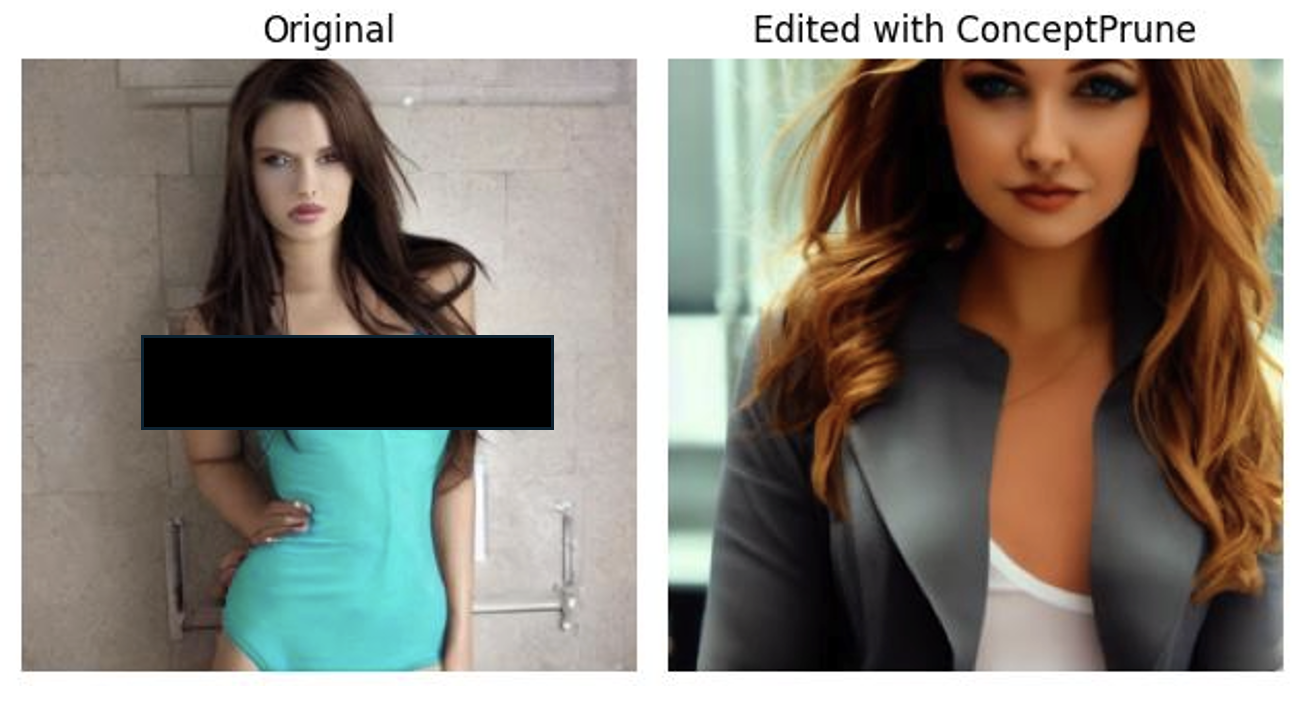} \\
\includegraphics[width=0.45\linewidth]{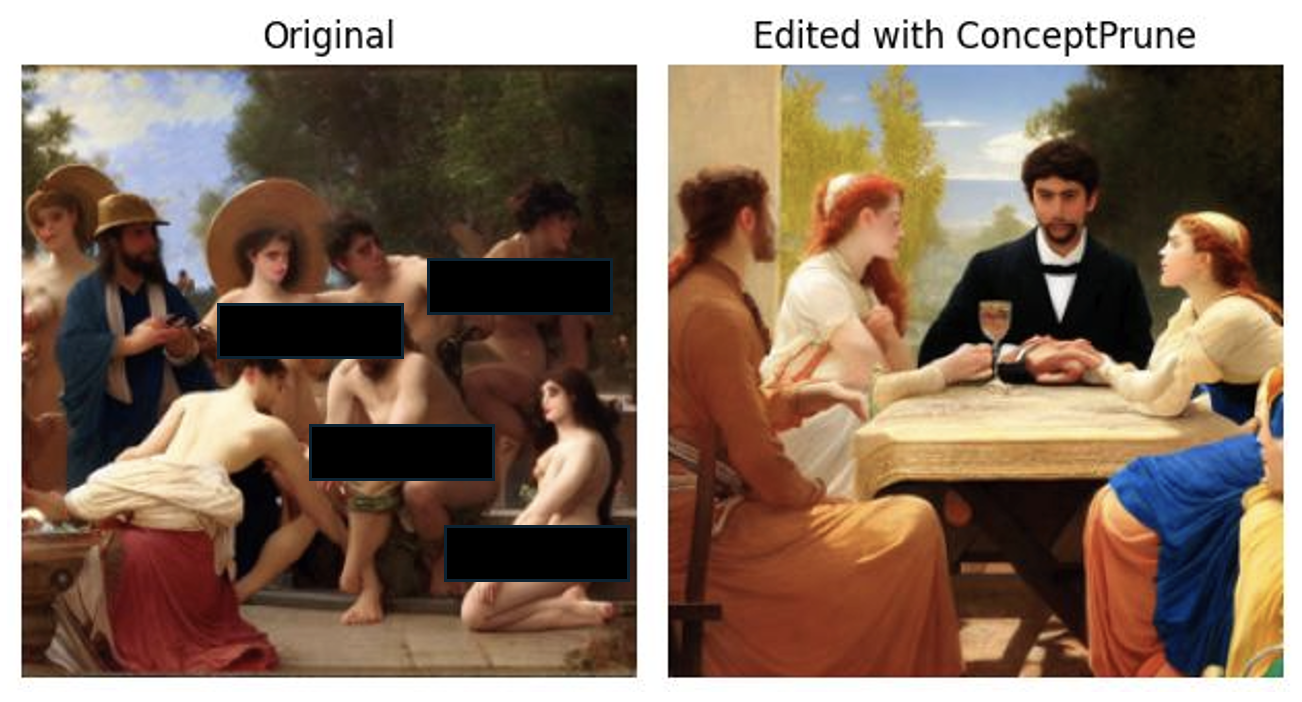} & \includegraphics[width=0.45\linewidth]{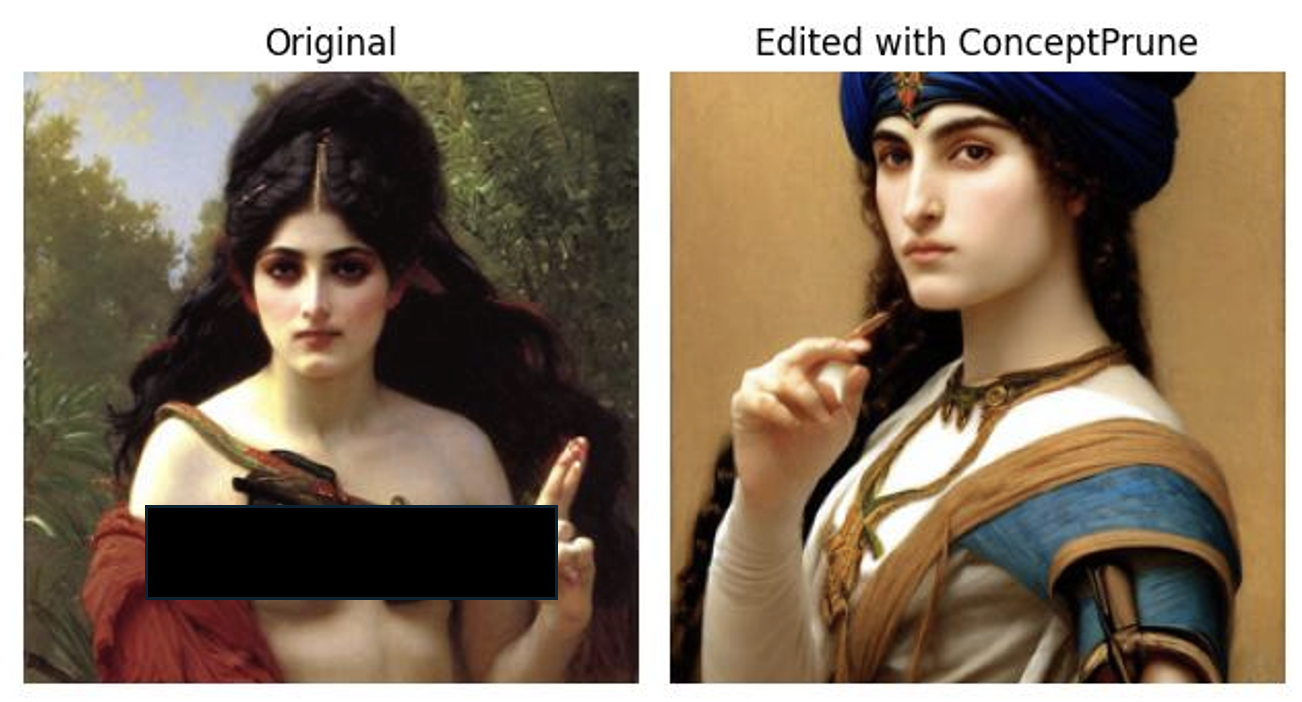} \\
\includegraphics[width=0.45\linewidth]{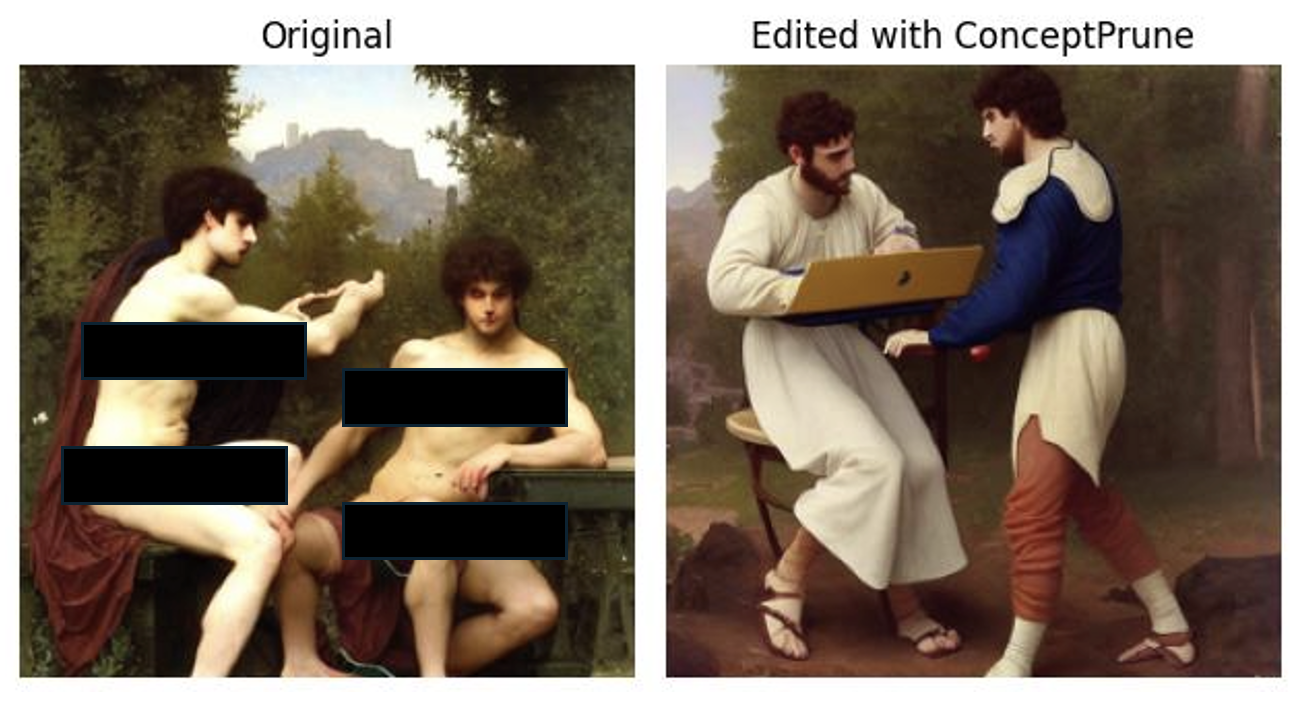} & \includegraphics[width=0.45\linewidth]{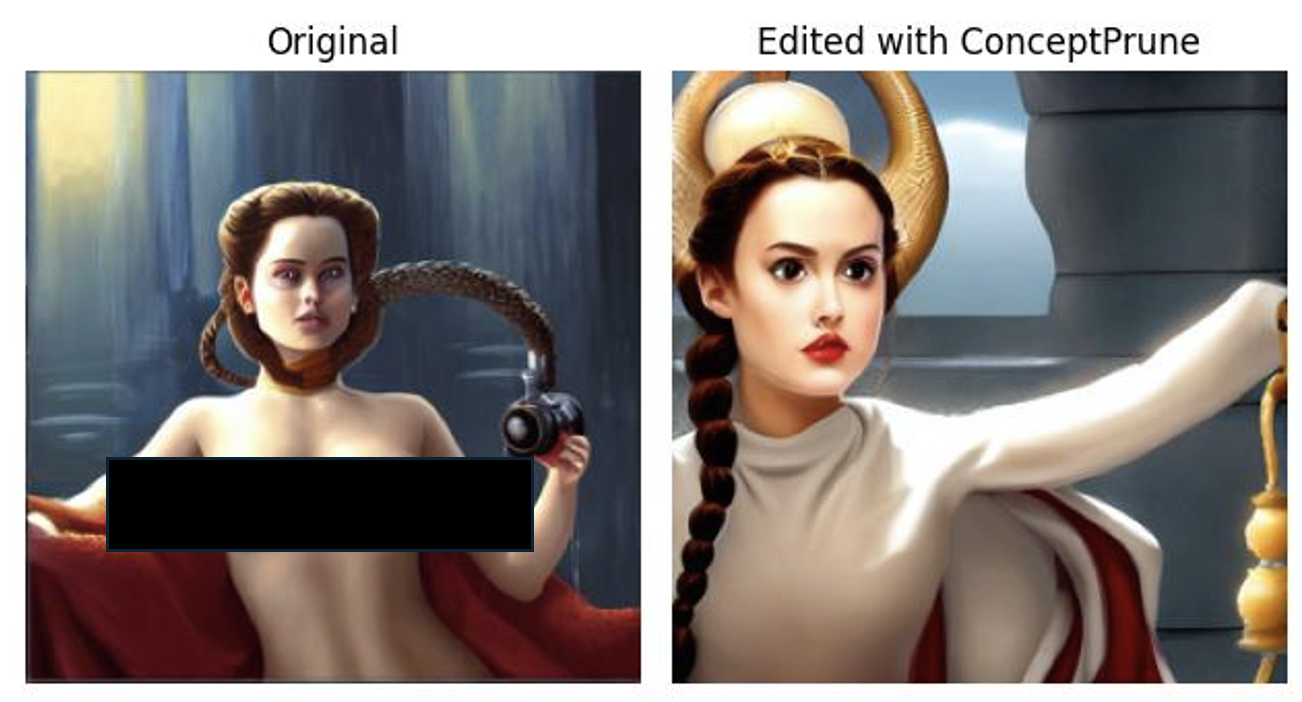} \\
\end{tabular}
    \captionof{figure}{Qualitative results for Nudity Erasure. We omit the prompts for safety. Images marked as "Original" correspond to images generated by pre-trained Stable Diffusion. Sensitive parts have been blacked out by the authors for the purpose of publication. We observe that ConceptPrune erases nudity while preserving other details and quality of the image.}
\label{fig:all-nsfw-images}
\end{table}

\begin{table}[]
    \centering
\begin{tabular}{cc}
\includegraphics[width=0.45\linewidth]{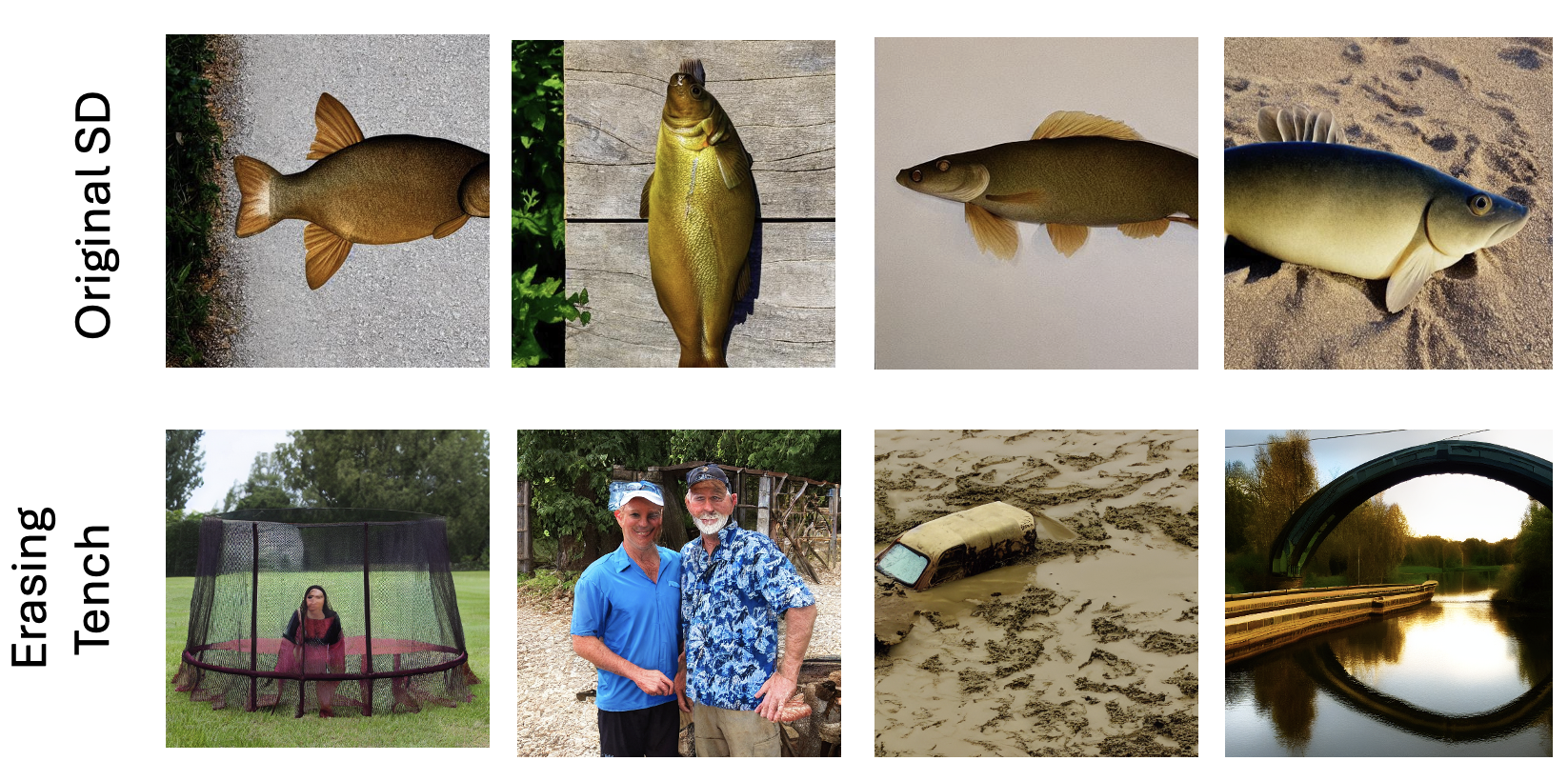} & \includegraphics[width=0.45\linewidth]{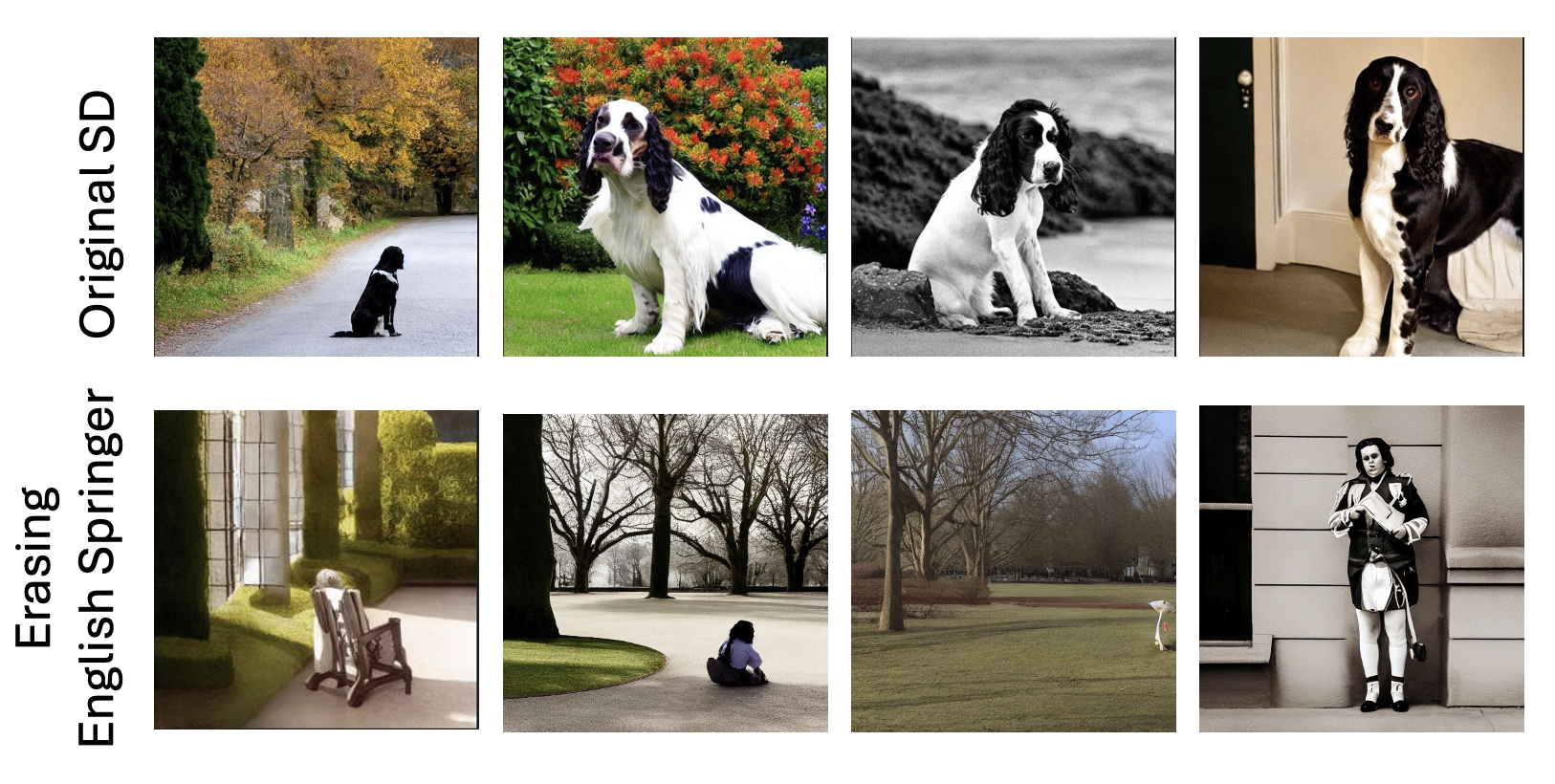}  \\
\includegraphics[width=0.45\linewidth]{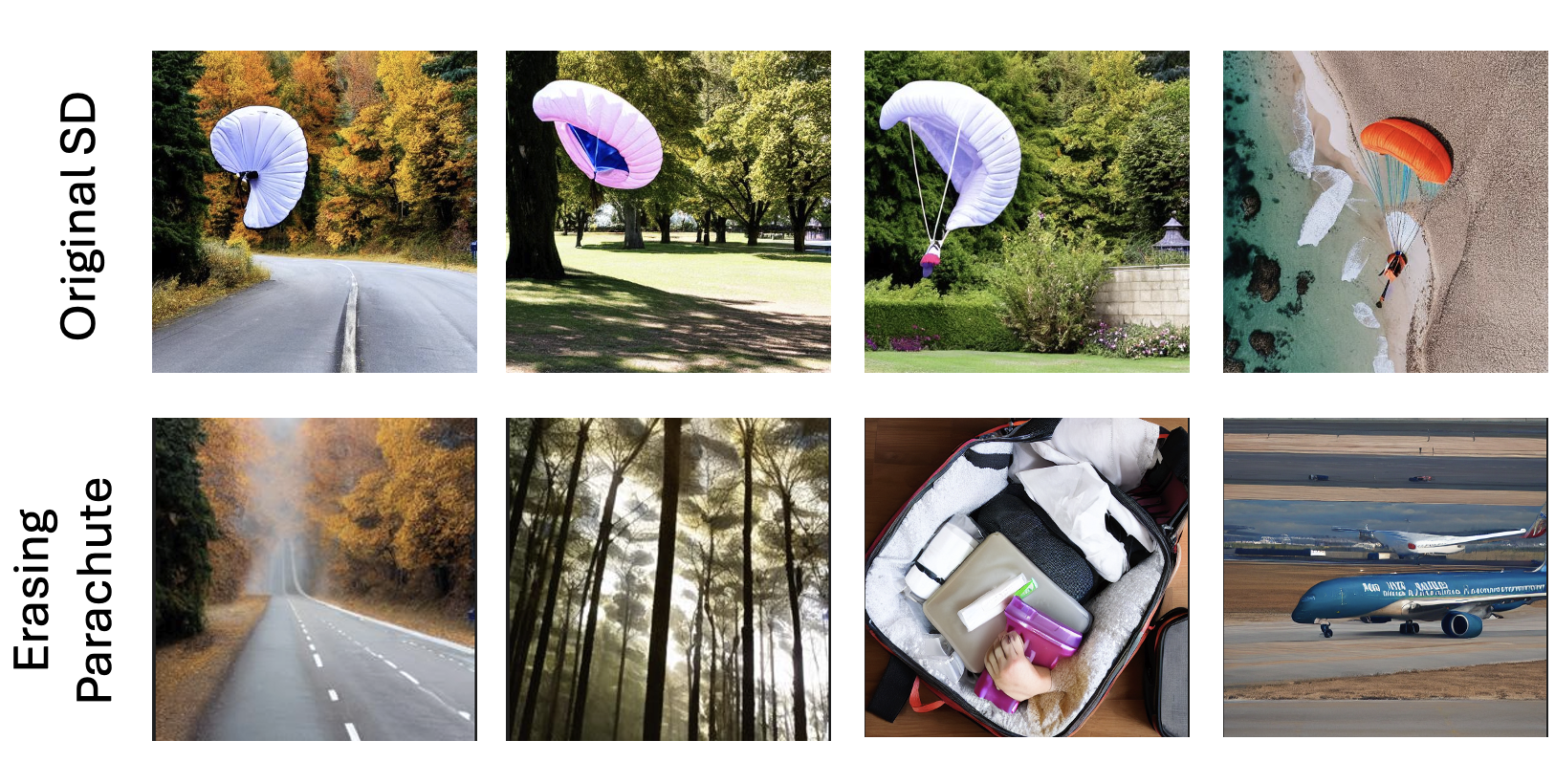} & \includegraphics[width=0.45\linewidth]{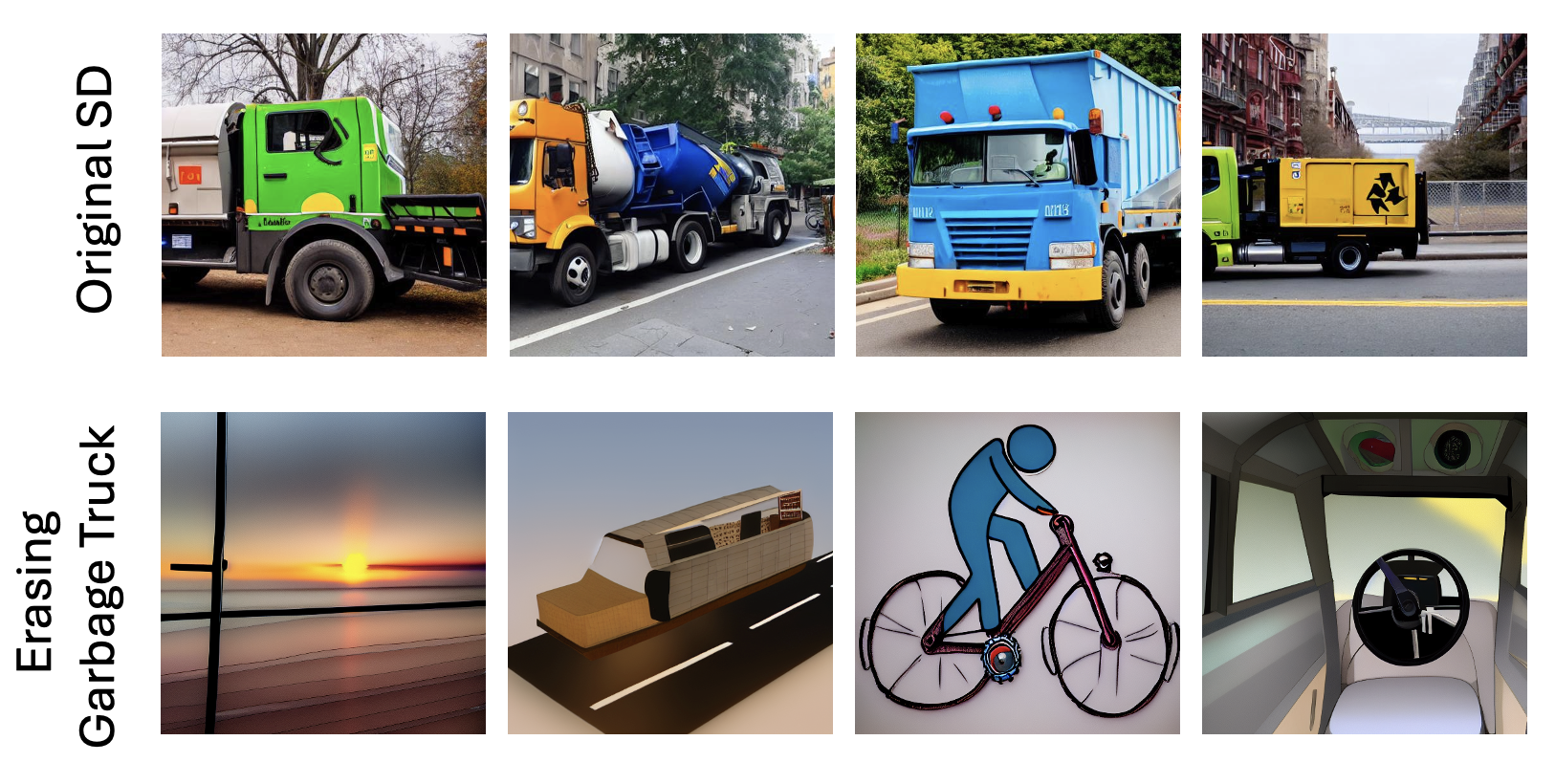} \\
\end{tabular}
    \captionof{figure}{Qualitative results for Object Erasure}
\label{fig:all-object-images}
\end{table}

\end{document}